\pdfoutput=1
\documentclass[pdflatex,sn-mathphys-num]{sn-jnl}% Math and Physical Sciences Numbered Reference Style
\usepackage{graphicx}%
\usepackage{multirow}%
\usepackage{amsmath,amssymb,amsfonts}%
\usepackage{amsthm}%
\usepackage{mathrsfs}%
\usepackage[title]{appendix}%
\usepackage{xcolor}%
\usepackage{textcomp}%
\usepackage{manyfoot}%
\usepackage{booktabs}%
\usepackage{algorithm}%
\usepackage{algorithmicx}%
\usepackage{algpseudocode}%
\usepackage{listings}%
\usepackage[normalem]{ulem}
\usepackage{longtable}% Supplementary Table 4

\definecolor{addcol}{HTML}{1A7F37}
\definecolor{chgcol}{HTML}{1F4FBF}
\definecolor{delcol}{HTML}{C0271D}
\newif\ifmarkup
\ifdefined\cleanversion\markupfalse\else\markuptrue\fi
\ifmarkup
  
  \newcommand{\chg}[1]{{\color{chgcol}#1}}
  \newcommand{\del}[1]{{\color{delcol}\sout{#1}}}
\else
  
  \newcommand{\chg}[1]{#1}
  \newcommand{\del}[1]{}
\fi

\theoremstyle{thmstyleone}%
\theoremstyle{thmstyletwo}%

\theoremstyle{thmstylethree}%

\begin{document}

\title[FloDR]{FloDR: an invertible dimensionality reduction method based on a normalising flow}


\author[1]{\fnm{Abdallah} \sur{Baraka}}\email{abdallah.baraka@wur.nl}

\author*[1]{\fnm{Daniel} \sur{Probst}}\email{daniel.probst@wur.nl}

\affil*[1]{\orgdiv{Bioinformatics Chair Group}, \orgname{Wageningen Unviersity}, \orgaddress{\street{Droevendaalsesteeg 1}, \city{Wageningen}, \postcode{6708 PB}, \country{the Netherlands}}}


\abstract{Maps of high-dimensional data are read for their neighbourhoods and global arrangements. Methods such as t-SNE and UMAP keep one at the expense of the other. We introduce FloDR, which fits the map as the first two coordinates of an invertible normalising flow trained on a neighbour objective and an ordinal term ranking input distances. Across seven data sets, raising the ordinal weight increased the rank correlation of input and map distances by 0.13 to 0.70, while neighbourhood recall moved by at most 0.015. On three of four single-cell atlases, FloDR kept more neighbours than UMAP and ranked distances better. On both versions of the OpenProblems dimensionality reduction task, FloDR had a higher mean score than every published method. Against unseen measurements, FloDR followed sampling geography in human genotypes better than seven of eight other maps, and clonal fate better than UMAP in two of three lineage data sets.}




\maketitle

\section{Introduction}\label{sec1}

A two-dimensional embedding of high-dimensional data is read as a map. Readers want to know which points are near each other and how groups of points lie relative to one another. While neighbour embeddings preserve local structure, they tend to distort global structure. Methods that prioritise global structure, such as principal component analysis (PCA) or classical multidimensional scaling (MDS), keep it but blur the fine-grained neighbourhoods that make localised clusters visible.

Both t-SNE and UMAP place points by pulling $k$-nearest-neighbour pairs together while pushing the others apart~\cite{Maaten2008VisualizingDU,McInnes2018UMAPUM}. Their attraction--repulsion trade-off can misrepresent global structure~\cite{Bohm2020AttractionRepulsionSI,Chari2021TheSA}. Several methods add a global term. PaCMAP adds mid-near pairs, TriMap optimises distance triplets, SQuadMDS adds a scale-free stochastic-quartet MDS term, and PCUMAP adds a correlation objective on measured distances to the UMAP loss~\cite{Wang2020UnderstandingHD,Amid2019TriMapLD,Lambert2022SQuadMDSAL,Sanju2025AdvancingDR}. DREAMS pulls each point of a t-SNE map towards its position on the first two principal components~\cite{kury2026}. Parametric UMAP, parametric t-SNE, ivis, and scvis replace the point optimisation with an encoder, enabling the embedding of new points in a single pass~\cite{Sainburg2020ParametricUE,Maaten2009LearningAP,Szubert2019StructurepreservingVO,Ding2017InterpretableDR}. The reverse is inverse projection, where iLAMP interpolates local affine maps, while NNInv, SSNP, and ShaRP train a decoder~\cite{Santos2012iLAMPEH,Espadoto2019DeepLI,Espadoto2021SelfsupervisedDR,Machado2023ShaRPSM}. In all these approaches, the forward and inverse maps are not exact inverses of each other.

Normalising flows are exact bijections with tractable Jacobians, and have been adapted to data on manifolds~\cite{Brehmer2020FlowsFS,Caterini2021RectangularFF,Flouris2023CanonicalNF}, to ordered latent dimensions~\cite{Bekasov2020OrderingDW}, building on nested dropout~\cite{Rippel2014LearningOR}, and to invertible reductions under isometry constraints~\cite{Li2020InvertibleML,Cramer2022NonlinearIM}, but none of them trains the leading coordinates with a neighbour-embedding objective.

Here, we present FloDR, a method that preserves both local and global structure in a two-dimensional embedding and keeps an exact inverse and a density, which we check on held-out cells. FloDR trains a bijection $f:\mathbb{R}^D\rightarrow\mathbb{R}^D$ built from affine coupling layers and reads its first two output coordinates as the embedding. We take inspiration from the work by Bekasov and Murray~\cite{Bekasov2020OrderingDW} and differ from it and the other flows above by driving the two leading coordinates with a neighbour-embedding objective and fitting the density on the frozen flow. An ordinal term in the objective, ranking pairs of input distances, gives the layout global structure. We test FloDR on bechmark data sets, single-cell atlases, and the OpenProblems dimensionality reduction task, and against measurements that no method sees.

\begin{figure}[t]
\centering
\includegraphics[width=\textwidth,height=0.86\textheight,keepaspectratio]{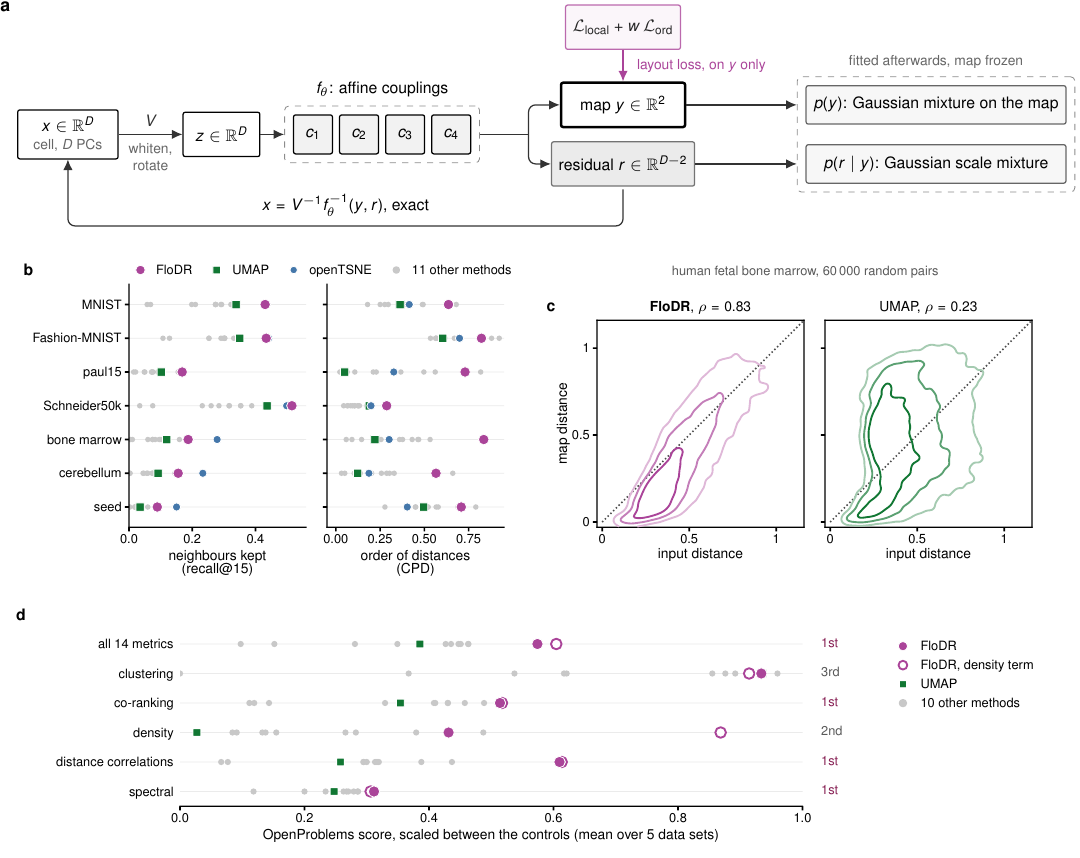}
\caption{\textbf{An invertible map, and what is kept by the layout.} \textbf{a}. The input is mapped by an exact affine bijection $V$ to whitened coordinates, rotated so that the layout starts from a plane fitted to the order of input distances, and is then passed through sequential affine coupling layers without mixing layers. The first two output coordinates are then read as the embedding $y$, and the remaining $D-2$ dimensions are retained as the residual $r$. The layout objective acts only on $y$. As the composition is a bijection, any pair $(y,r)$ maps back to an input up to a floating point round-off. With the frozen map after training, we fit a Gaussian mixture $p(y)$ to the map and a Gaussian scale mixture $p(r\mid y)$ to the residual. With the log-determinant of the Jacobian they give a density of $x$.\textbf{b}. Kept neighbourhoods (recall@15) and order of distances (Spearman correlation of pairwise distances, CPD) on the four benchmark data sets and the three atlases, means over three seeds. FloDR ($w=2$), UMAP, and openTSNE are highlighted, the eleven other methods of Fig.~\ref{fig:frontier} are shown in grey. \textbf{c}. Map distances against input distance for $60\,000$ random pairs of cells of the human fetal bone marrow atlas, FloDR compared to UMAP. Lines indicate the smallest regions that hold 50, 80, and 95\% of the pairs. $\rho$ is the Spearman correlation of the pairs shown, with axes scaled to the 99th percentile. \textbf{d}. Scores on the dimensionality reduction task of OpenProblems. Each metric is scaled per data set between the worst and the best of three control methods and then averaged over the data sets shown in Table~\ref{tab:openproblems}. A missing value counts as zero. 14 metrics are divided into 5 families: co-ranking (7), distance correlations (3), clustering (2), density (1), and spectral distance correlation (1).}
\label{fig:arch}
\end{figure}

\section{Results}
\label{sec:results}

\subsection{FloDR reads a layout from an invertible map and keeps the residual}

FloDR embeds a data set $X=\{x_1,\dots,x_n\}\subset\mathbb{R}^D$ by training a bijection $f_\theta:\mathbb{R}^D\rightarrow\mathbb{R}^D$ and using the first two output coordinates $y\in\mathbb{R}^2$ for the visualisation and keeping the residual $r\in\mathbb{R}^{D-2}$ (Fig.~\ref{fig:arch}). As $f_\theta$ is a bijection, the embedding, its inverse, and a density $\log p_\theta(x)$ are properties of a single object. The objective only acts on the two plotted coordinates and combines a contrastive~\cite{damrich2023} loss on UMAP's fuzzy $k$-nearest-neighbour graph, which sets each edge against random negatives of its ends, with an ordinal hinge that uses only the sign of measured distance differences. Therefore, any dissimilarity can serve as the input metric. The layout starts from a plane fitted to the order of input distances. Unseen points are embedded with a single forward pass. This costs local structure, which a second step restores by optimising the new positions. It kept more neighbours of held-out points than the transform of UMAP and fewer than that of openTSNE on all seven evaluated data sets, at a higher rank correlation of distances than either (Methods and Extended Data Fig.~\ref{fig:heldout_frontier}). Once trained, a density over the embedding, $p(y)$, and a density of the residual given the position, $p(r|y)$, are fitted, and drawing a residual from $p(r|y)$ and running it backwards through the map gives an input that the map places exactly at $y$ (Methods). As this density is only exact under the model, we check that held-out cells rank uniformly among its exact draws. On the human fetal bone morrow atlas, the density on the map passed this check on all three fits, but $p(r|y)$ failed it, as did two alternative forms (Methods and Extended Data Fig.~\ref{fig:density}). Because a position alone leaves the residual open, we decode the position from the mean of the 15 nearest training points on the map, moved to a nearby input that the map places exactly at the position. Held-out points that are placed by the optimised step, which, like the transform of UMAP, uses their nearest training points in the input, were decoded in this way with a lower error than by the inverse transform of UMAP and the decoder of Parametric UMAP on all 24 splits of six data sets (Extended Data Fig. \ref{fig:decoding}). Placed by the forward pass, which, like the encoder of Parametric UMAP, uses no neighbours,they were decoded less accurately than by Parametric UMAP on 19 of the splits. The same mean was less accurate on UMAP's than on FloDR's map on every split, but on the images it matched or beat the exact decode, as exactness raised the error by at most 3~\% except on Fashion-MNIST, from 0.30 to 0.33 (Methods and Supplementary Note 1).

\subsection{An ordinal term in the objective adds global structure at little cost in local structure}

We compared FloDR against neighbour embedding methods and PCA as a linear control on four data sets (images of MNIST and Fashion-MNIST, single-cell transcriptomes of paul15, and reaction fingerprints of Schneider50k). Scoring every method on an identical input matrix, we report the neighbourhood recall at $k=15$ for local structure and the Spearman correlation of pairwise distances (CPD) for global structure. CPD is only scored on pairs that the ordinal term never samples in training (Methods).

\begin{figure}[t]
\centering
\includegraphics[width=0.97\textwidth,height=0.86\textheight,keepaspectratio]{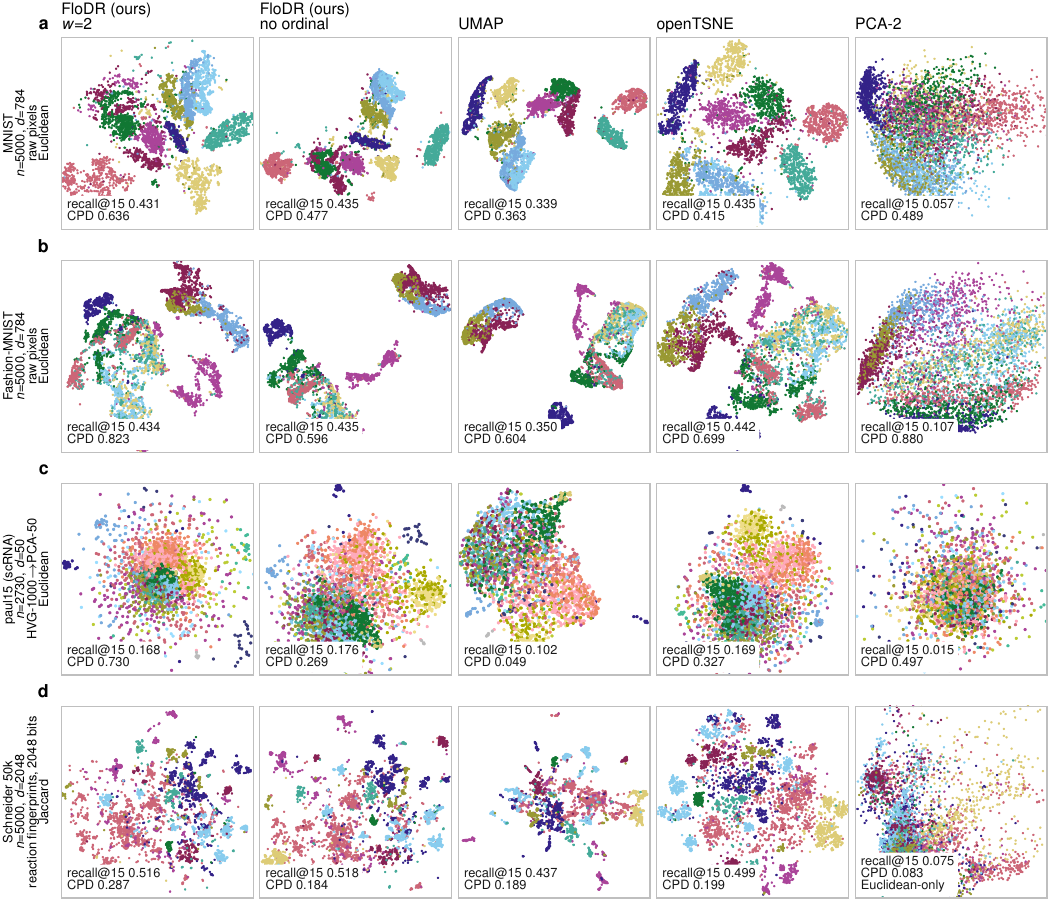}
\caption{\textbf{Layouts of the four benchmark data sets.} One data set per row and one method per column. Each row label gives the sample size $n$, the input dimension $d$, the representation, and the metric the data set is scored in. Points are coloured by class label. Annotations give recall at $k=15$ and CPD, means over three seeds. The first column is FloDR at defaults, $w=2$, and the second FloDR without the ordinal term ($w=0$). PCA on Schneider50k was fitted on Euclidean distance but scored on the Jaccard metric that it cannot consume. Extended Data Fig.~\ref{fig:comparison_appendix} shows the remaining baselines.}
\label{fig:comparison}
\end{figure}

Global structure comes from the ordinal term with weight $w$. Starting from a random plane instead of the fitted one lowers the CPD on paul15 by only 0.009 (Methods). Removing the term (Fig.~\ref{fig:comparison}, second column) changes recall by at most 0.008 on the four data sets, while the term raises the CPD from $0.477$ to $0.636$, $0.596$ to $0.823$, $0.269$ to $0.730$, and $0.184$ to $0.287$. Further baselines are shown in Extended Data Fig.~\ref{fig:comparison_appendix}, and all are scored in Extended Data Table~\ref{tab:main-comparison}. With the exception of training for 800 instead of $6\,400$ steps, changing the other components of FloDR one at a time moved recall by at most $0.004$ (Extended Data Table~\ref{tab:ablations}). FloDR with $w=2$ (the default) performs better than UMAP on both axes on all four data sets. openTSNE retains more local structure on three of the four at a substantial cost in global structure, and on Schneider50k, FloDR retains more neighbours and ranks Jaccard distances better than every other method. Of the others, only UMAP, openTSNE and DREAMS used Jaccard distance (Methods).

\begin{figure}[t]
\centering
\includegraphics[width=0.97\textwidth,height=0.86\textheight,keepaspectratio]{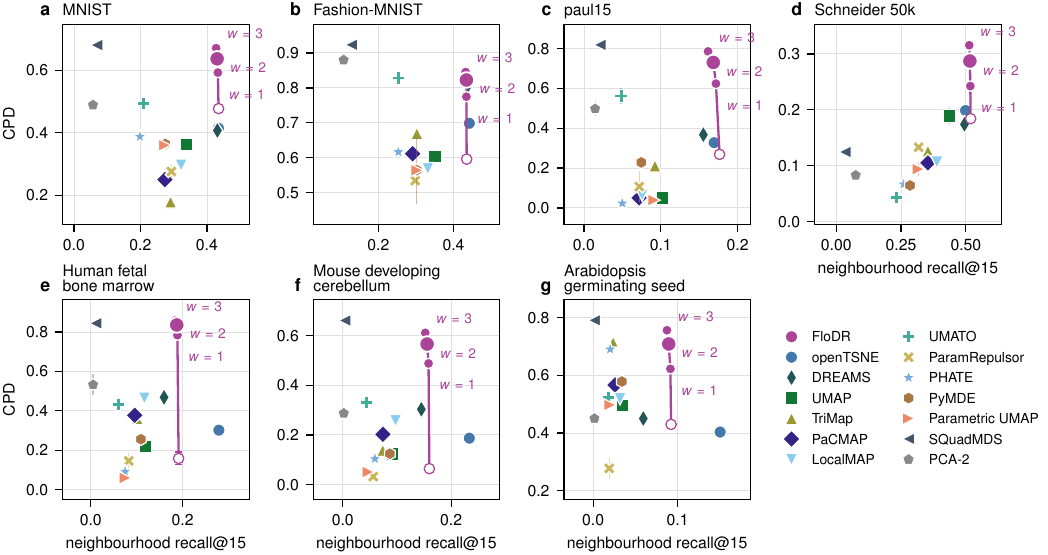}
\caption{\textbf{Local-to-global trade-off across seven data sets.} Neighbourhood recall is at $k=15$ against CPD. The upper-right corner is better on both axes. The magenta curve is the FloDR sweep over the ordinal weight $w\in\{0,1,2,3\}$, all baselines contribute a single point. Points are means and error bars standard deviations over three seeds. The open circle is FloDR at $w=0$.}
\label{fig:frontier}
\end{figure}

Sweeping the ordinal weight $w\in\{0,1,2,3\}$ barely moves local recall ($\leq0.015$) across seven data sets (MNIST, Fashion-MNIST, paul15, Schneider50k, human fetal bone marrow, developing mouse cerebellum, and germinating \textit{Arabidopsis} seed), but increases CPD by $0.13$ to $0.70$ from $w=0$ to $w=3$, so that the sweep is an almost vertical line in Fig.~\ref{fig:frontier}. On the three single-cell atlases, FloDR keeps more neighbours than UMAP (Fig.~\ref{fig:arch}b), with a recall of $0.187$, $0.156$ and $0.089$ against $0.119$, $0.092$ and $0.035$, at a CPD of $0.836$, $0.566$, and $0.709$ against $0.220$, $0.123$, and $0.495$. There, openTSNE keeps substantially more neighbours than all other methods.

FloDR leads on neither axis on three of the four benchmark data sets, as openTSNE keeps more neighbours and SQuadMDS, UMATO, or PCA rank distances better. On these three, FloDR's recall is within 0.008 of openTSNE's, while its CPD is higher by 0.12 to 0.40 (Fig.~\ref{fig:comparison}). SQuadMDS, which we ran as quartet loss alone (Methods), ranks distances better than FloDR on every atlas ($0.844$, $0.661$, and $0.791$) but keeps almost no neighbours (recall of $0.011$, $0.0035$, and $0.0014$). Like FloDR, DREAMS aims to preserve both local and global structure. Against its default regularisation strength, $\lambda=0.15$, FloDR ranked distances better on all seven data sets, by 0.014 to 0.37, with recall within 0.005 on MNIST and Fashion-MNIST, and higher on the other five (Supplementary Table 1). None of the six values for $\lambda$, from $\lambda=0$ (t-SNE), to $\lambda=0.8$, was ahead of FloDR on both axes on any data set, and only on Fashion-MNIST did a stronger regularisation rank distances better, at a lower recall. Of 16 further settings of UMAP, openTSNE, and PaCMAP, none kept more neighbours than FloDR, and only openTSNE at an exaggeration of 8 or 12 ranked distances better, on the seed atlas and at a recall below 0.02 (Supplementary Table 2 and Methods).

\subsection{FloDR preserves global structure in single-cell atlases}

\begin{figure}[p]
\centering
\includegraphics[width=\textwidth,height=0.86\textheight,keepaspectratio]{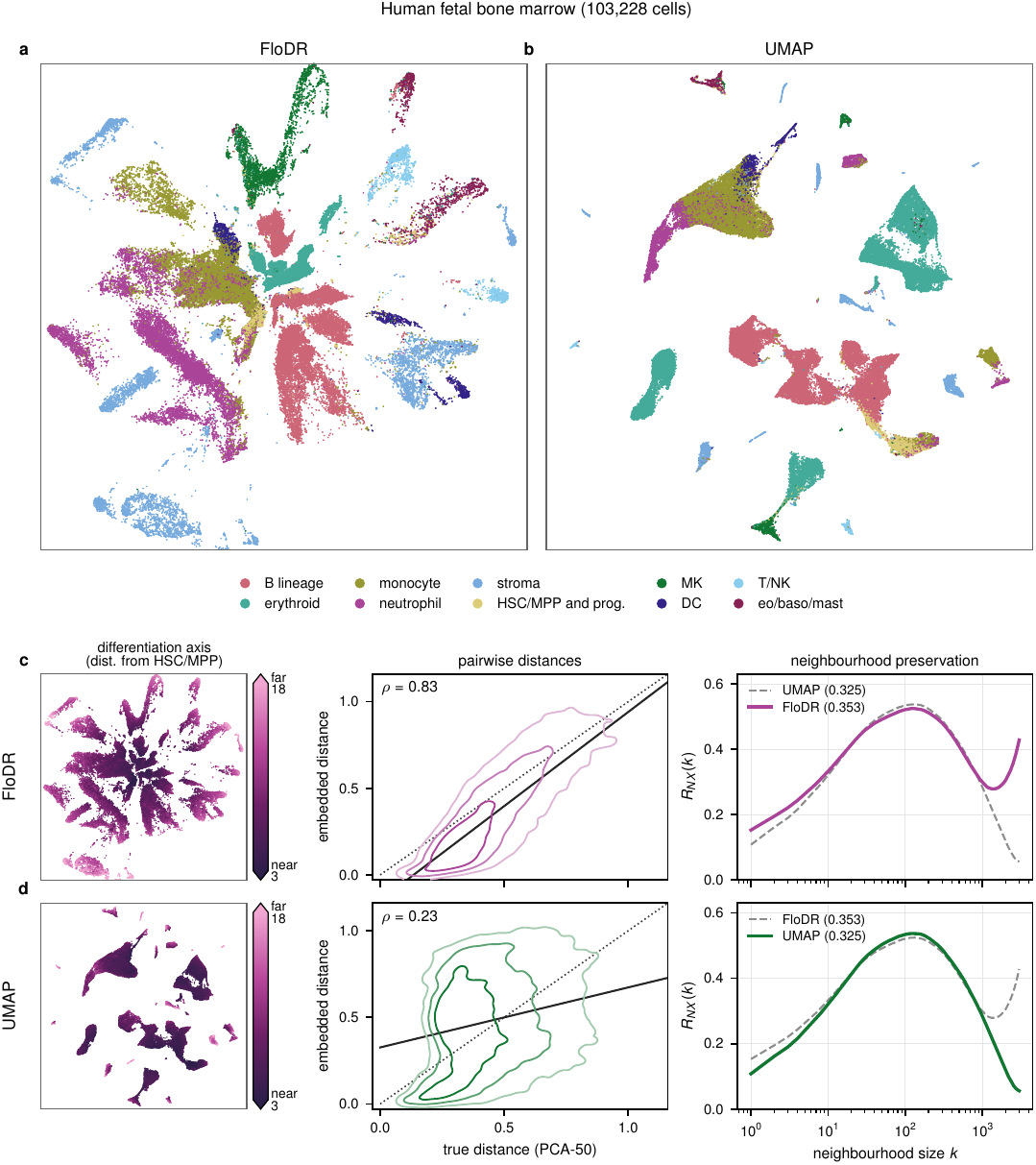}
\caption{\textbf{FloDR and UMAP embeddings of the human fetal bone marrow atlas} ($n=103\,228$ cells, 10 annotated cell types). The top row (\textbf{a} FloDR, \textbf{b} UMAP) shows the two-dimensional embeddings coloured by cell type, with orders shuffled to avoid overplotting. Below, each method is analysed in a row (\textbf{c} FloDR, \textbf{d} UMAP). The first column colours the same embedding by distance from the progenitor centroid in PCA-50 space. The second column is a Shepard plot of $60\,000$ random pairs, with lines around the smallest regions that hold 50, 80 and 95\% of the pairs, where the dotted line represents the diagonal of the rescaled axes and the solid line the least-squares fit. $\rho$ is the Spearman correlation of the underlying distances. The third column shows the co-ranking curve $R_{NX}(k)$ against neighbourhood size on a logarithmic axis, computed on a common subsample of $6\,000$ cells, up to $k=n/2$, and the numbers in its legend are the log-weighted areas under the curves. Extended Data Figs.~\ref{fig:cerebellum} and~\ref{fig:plant} show the same plots for the mouse developing cerebellum and \textit{Arabidopsis} germinating seed atlases.}
\label{fig:bmarrow}
\end{figure}

On the human fetal bone marrow atlas, the Spearman correlation of $60\,000$ random pairs is $0.83$ for FloDR and only $0.23$ for UMAP (Fig.~\ref{fig:arch}c), whose map distances stay close to flat over most of the range of input distances (one fit each, Figs.~\ref{fig:arch}c and \ref{fig:bmarrow}). The gain  global structure costs little local structure, as the co-ranking curves of Fig.~\ref{fig:bmarrow} show FloDR ahead of UMAP at small and large neighbourhood sizes, and at most $0.014$ behind it between about $20$ and $800$ neighbours. The cerebellum and seed atlases are shown in Extended Data Figs.~\ref{fig:cerebellum} and~\ref{fig:plant}. The cell types also sit where the input places them (Fig.~\ref{fig:bmarrow}a,b). Over three fits, the median distances between the ten annotated types kept their input order the FloDR map, with a Spearman correlation of 0.82 to 0.83, against 0.25 to 0.29 for UMAP. Among the closest third of the pairs of the 56 finer types it was 0.84 to 0.87 versus 0.21 to 0.23. The distance of each cell from the progenitors followed that of the input with 0.78 to 0.80 versus 0.34 to 0.37 (Fig.~\ref{fig:bmarrow}c,d). As in the 50 principal components, the erythroid cells, the progenitors, and the monocytes sit at the centre of the FloDR map, the plasmacytoid dendritic cells apart from the monocytes, and the stroma at the far end. FloDR keeps the erythroid stages in one piece (93 to 100~\% each) and the megakaryocytes as near to them as in the input. UMAP places the B lineage at the centre, joins the neutrophils to the monocytes and sets the erythroid and megakaryocyte cells far apart. In every fit, it draws the megakaryocyte-erythroid progenitors and 62 to 63~\% of the early erythroid cells as a bridge to early megakaryocytes. FloDR draws this bridge in one of three fits and otherwise places most of these progenitors with the erythroid cells, which outnumber megakaryocytes two to one among their input neighbours. Both maps split off cells of a single donor in at least two of three fits, as the input is not corrected for donor or sequencing chemistry. By distance from the progenitors, UMAP orders the maturation stages of each lineage more strictly than the input, while FloDR stays at least as near the order of the input. FloDR draws eosinophils, basophils, and mast cells closer to T and NK cells than the input does. These two annotated types, the smallest, lie almost equally far from each of the other eight, so a plane puts them in one region. For this atlas, the map of FloDR therefore agrees better with the input on how the cell types and lineages related.

\subsection{A fast variant fits a million cells in about two minutes on a GPU at a lower recall}
A fast variant of FloDR runs a quarter of the $6\,400$ training steps on two of the four coupling layers. On a developing zebrafish atlas~\cite{Duran2025}, it fitted $10^6$ cells in 128~s, against 612~s for the default and 403~s for UMAP on one CPU thread (Extended Data Fig. \ref{fig:scaling} and Methods). The vast variant's recall was within $0.003$ of the default's on four benchmark data sets (Extended Data Table~\ref{tab:main-comparison}) and lower by $0.017$ to $0.021$ on the three atlases, with CPD within $0.022$ and both scores above those of UMAP on all seven. On all cells of the zebrafish atlas, its recall was $0.0175$, below the $0.0207$ of the default and the $0.0215$ of UMAP, at a CPD of $0.724$ versus $0.731$ and $0.342$. It also placed new points less precisely (Methods).

\subsection{FloDR has the highest mean score on the OpenProblems benchmark}
The dimensionality reduction task of OpenProblems, a community-driven benchmark for single-cell analysis~\cite{Luecken2025}, gives every method the normalised counts of a data set and scores the two-dimensional embedding on 14 metrics in five families. We compared FloDR with the published scores of other methods on the five of its 14 data sets in which every metric can be scaled (Methods), namely, mouse myeloid progenitors, mouse haematopoietic stem and progenitor cells (HSPC), human pancreas, adult mouse visual cortex, and zebrafish embryos. Fitted at its defaults on 500 principal components, FloDR had the highest mean score with $0.574$ against $0.463$ for NeuralEE and PHATE, the next best, and $0.385$ for UMAP (Table~\ref{tab:openproblems}). It ranked first on co-ranking, distance correlations, and spectral distance correlation, second on density behind densMAP, and third on clustering behind ivis and UMAP (Fig.~\ref{fig:arch}d and Extended Data Table~\ref{tab:openproblems-metrics}). With the density term of densMAP, the density score of FloDR rose from $0.431$ to $0.869$ and its mean from $0.574$ to $0.605$, and at two decimals only the adjusted Rand index and the local co-ranking quality $Q_\mathrm{loc}$ fell. On the three data sets on which densMAP has a density score, FloDR with the density term scored $0.847$ on density, against $0.812$ for densMAP.

\begin{table}[h]
\caption{\textbf{Scores on the dimensionality reduction task of OpenProblems.} Each of the 14 metrics is scaled per data set between the worst and the best of three control methods, clipped to $[0,1]$, and averaged over metrics and data sets. A missing value counts as zero, also within a data set, as on the leaderboard of the benchmark, and a dash marks a data set lacking published scores for a method. The last column averages only the available values per data set and then over the data sets with values. It does not compare like with like, as SIMLR has only values on mouse myeloid. On only the values of each method, FloDR leads all but SIMLR, which is ahead by $0.026$ on its mouse myeloid values, and densMAP comes closest of the others, with $0.065$ behind. FloDR was scored on every metric (Methods), while the other rows are the published scores of the task. The data sets are mouse myeloid progenitors (Myeloid), mouse HSPC, human pancreas, adult mouse visual cortex (Cortex), and zebrafish embryos.}
\label{tab:openproblems}
{\setlength{\tabcolsep}{4pt}
\begin{tabular}{@{}lccccccc@{}}
\toprule
 & & & & & & \multicolumn{2}{c}{Mean} \\
\cmidrule(l){7-8}
Method & Myeloid & HSPC & Pancreas & Cortex & Zebrafish & All & Completed \\
\midrule
FloDR & 0.608 & 0.444 & 0.403 & 0.872 & 0.544 & 0.574 & 0.574 \\
FloDR, density term & 0.690 & 0.457 & 0.422 & 0.892 & 0.562 & 0.605 & 0.605 \\
\midrule
NeuralEE & 0.615 & 0.261 & 0.323 & 0.731 & 0.387 & 0.463 & 0.463 \\
PHATE & 0.549 & 0.267 & 0.378 & 0.713 & 0.407 & 0.463 & 0.467 \\
ivis & 0.559 & 0.246 & 0.348 & 0.705 & 0.397 & 0.451 & 0.451 \\
t-SNE & 0.551 & 0.252 & 0.284 & 0.699 & 0.451 & 0.447 & 0.456 \\
PyMDE & 0.586 & 0.264 & 0.168 & 0.709 & 0.450 & 0.435 & 0.469 \\
PCA & 0.473 & 0.222 & 0.281 & 0.759 & 0.398 & 0.427 & 0.433 \\
UMAP & 0.470 & 0.206 & 0.305 & 0.702 & 0.245 & 0.385 & 0.455 \\
LMDS & 0.526 & 0.223 & 0.272 & 0.726 & -- & 0.349 & 0.455 \\
densMAP & 0.647 & 0.283 & 0.188 & -- & 0.287 & 0.281 & 0.486 \\
Diffusion map & 0.411 & 0.162 & 0.185 & -- & -- & 0.152 & 0.253 \\
SIMLR & 0.488 & -- & -- & -- & -- & 0.098 & 0.569 \\
\botrule
\end{tabular}}
\end{table}

FloDR scores highest on four of the five data sets, leading by $0.161$ on HSPC, $0.113$ on cortex, $0.094$ on zebrafish, and $0.025$ on pancreas. On mouse myeloid, it scores $0.608$, behind densMAP ($0.647$) and NeuralEE ($0.615$), and ahead of both with the density term ($0.690$). On the first version of the task, shown on the OpenProblems website (four data sets, 10 metrics), FloDR would also rank first among 23 method variants, scoring $0.400$ against $0.355$ for densMAP (Supplementary Table 3).

\subsection{FloDR follows sampling geography, and clonal fate in two of three lineage data sets, more closely than UMAP}
\label{sec:following}
\begin{figure}[p]
\centering
\includegraphics[width=\textwidth,height=0.86\textheight,keepaspectratio]{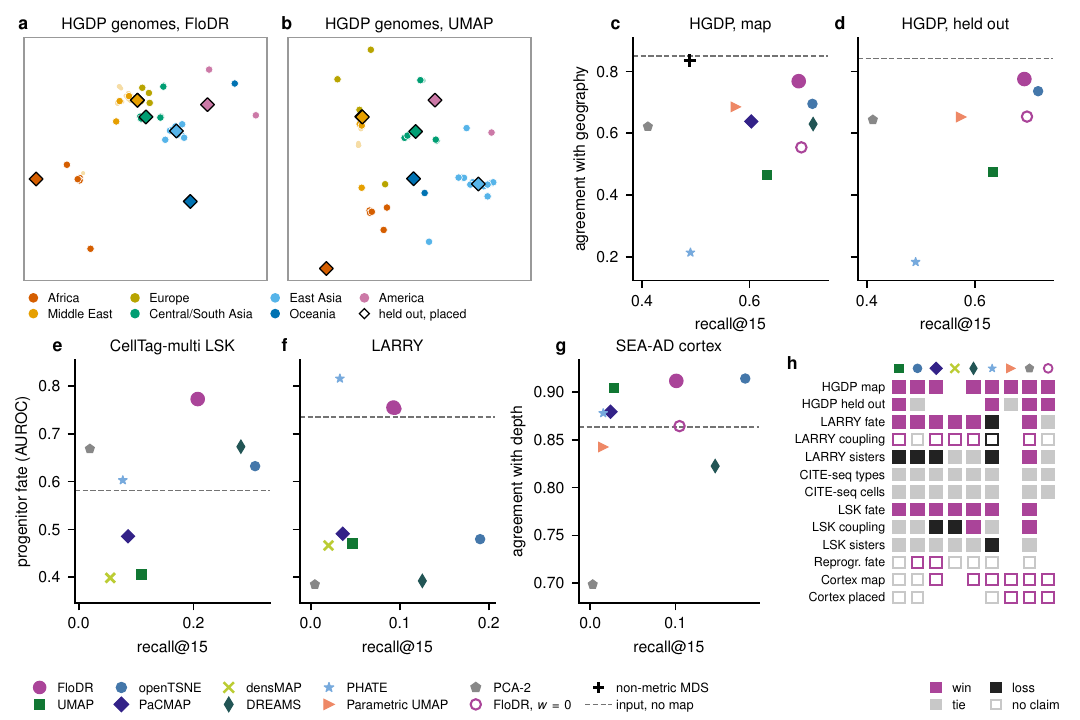}
\caption{\textbf{Maps against measurements that no method sees.} \textbf{a}, \textbf{b}. FloDR and UMAP maps (seed 0) of the genotypes of 795 individuals from 47 populations of the HGDP-CEPH panel, coloured by region. Dots are population medians and diamonds the seven held-out populations placed on the map. \textbf{c}. Agreement with geography (rank correlation of map distances between populations with distances between their sampling sites) versus neighbourhood recall ($k=15$, means over five seeds). \textbf{d}. The same for the held-out populations. \textbf{e}, \textbf{f}. AUROC for ranking of barcoded progenitors by the fate of most of their clone, on the LSK arm (lineage-negative, Sca-1-positive, c-Kit-positive progenitors) of CellTag-multi (647 progenitors), and on LARRY (828 progenitors). \textbf{g} Rank correlation of the map distances between ten bins of cortical depth with their separation for the 19 donors. \textbf{h} Comparison of FloDR against each other map (columns) on each readout. Open squares mark readouts that do not support a claim, as the input did not exceed UMAP by $0.03$, their outline nonetheless shows the result. The test of placed progenitors supported no claim and is not drawn.  For readouts and rules, see Methods and Supplementary Note 2.}
\label{fig:measurements}
\end{figure}

As CPD measures what the ordinal term is trained to keep, we also ran seven tests against measurements that no method sees. We decided wins, ties, and losses by a bootstrap rule (Fig.~\ref{fig:measurements}, Methods, and Supplementary Table 4). On genotypes of 47 human populations~\cite{Bergstrm2020}, the rank correlation of map distances with the distances between sampling sites~\cite{Ramachandran2005} was $0.768$ for FloDR, $0.695$ for openTSNE, and $0.465$ for UMAP, against $0.850$ for the input. FloDR won against all seven other maps of the test. Its placement of held-out populations won against UMAP and tied with openTSNE and Parametric UMAP. A non-metric MDS~\cite{Kruskal1964MultidimensionalSB} as a control agreed better with geography ($0.835$) but kept fewer neighbours. On the CellTag-multi (LSK arm)~\cite{Jindal2023}, FloDR ranked cells by their fate of their clone with an AUROC of $0.773$, against $0.581$ in the input and $0.405$ for UMAP, winning again against all other maps. However, two maps reproduced the coupling of cell types within clones better, while PHATE kept sister cells closer. On LARRY~\cite{Weinreb2020}, FloDR performs better than openTSNE and UMAP with an AUROC of $0.755$ against $0.480$ and $0.471$, only losing against PHATE ($0.816$) on fate. Tests on CITE-seq protein distances~\cite{luecken2021sandbox} resulted in ties, and tests on reprogramming, placed progenitors, and cortical depth~\cite{Gabitto2024} supported no claim, as their input did not exceed UMAP by $0.03$. On the five of these data that were scored for neighbours and distance ranks, FloDR again kept more neighbours than UMAP and ranked distances better, while openTSNE and DREAMS kept more neighbours (Supplementary Table 5).

\section{Discussion}
\label{sec:discussion}

With FloDR, we have shown that an ordinal term adds global structure to a neighbour embedding at little cost in local structure. On three single-cell atlases, FloDR kept more of the 15 nearest neighbours than UMAP while ranking distances better, and it had the highest mean score in both versions of the OpenProblems dimensionality reduction task. On measurements that no method sees, it followed sampling geography in human genotypes more closely than seven of eight other maps.

The ordinal term uses only the sign of the difference between two input distances, so it asks the map to keep the order of distances and leaves their values free. This may be why it costs the neighbour objective little. On the OpenProblems benchmark, whose metrics we did not define, FloDR leads most clearly on distance correlations, which are closest to what the ordinal term ranks. The objective sets the layout. Changing the number of coupling layers moved recall by at most $0.002$ and CPD by at most $0.007$, far less than removing the ordinal term. The flow makes this layout part of a bijection with an exact inverse and density. FloDR has limits. On a GPU, the default fit took 30 to 119~s on the four benchmark data sets, against 5 to 20s for UMAP and openTSNE on the CPU (Extended Data Table~\ref{tab:main-comparison}), and 1.5 to 4.5 times as long as UMAP on the zebrafish atlas (Extended Data Fig.~\ref{fig:scaling}. On all cells of this atlas, the default kept slightly fewer neighbours than UMAP. On sixof the seven data sets, openTSNE keeps more neighbours. On unseen measurements, a non-metric MDS agreed better with geography and PHATE followed fate more closely on LARRY. Finally, the density is only exact under the model.

FloDR is available as an open-source Python package with the scikit-learn interface of UMAP and can take its place in existing analyses.

\section{Methods}
\label{sec:methods}

\subsection{The Map}
\label{sec:the_map}
The map is a composition of $L=4$ affine coupling layers, specifically RealNVP by Dinh et al.~\cite{Dinh2016DensityEU}, without mixing layers between them. Layer $\ell$ splits the coordinates by a fixed mask $m_\ell \in \{0,1\}^D$, where layers 2 and 4 update the two map coordinates given the residual, and layers 1 and 3 each update one alternating half of the residual given the map coordinates and the other half, and then applies
\begin{equation}
u' = m_\ell \odot u + (1 - m_\ell) \odot \bigl(u \odot e^{s_\ell(u_m)} + t_\ell(u_m)\bigr),
\qquad u_m = m_\ell\odot u,
\label{eq:coupling}
\end{equation}
where $s_\ell$ and $t_\ell$ are the outputs of a shared MLP with two hidden layers of width 256 and tanh activations. The final layer of each MLP is initialised to zero, as in Glow~\cite{Kingma2018GlowGF}, so that every coupling starts as identity and the map starts as the input parameterisation described below. As in RealNVP, the log scale is bounded as $s_\ell=g_\ell\tanh(\tilde{s}_\ell)$, where $g_\ell$ is a learnable gate with one entry per coordinate. This bound keeps the stretch factors below $e^{g_\ell}$ to avoid extreme scales but still allows growth depending on the data. The gate itself is bounded by $g_\ell=g_\mathrm{max}\tanh(\tilde{g}_\ell)$, where $\tilde{g}_\ell$ is a free parameter. We set $g_\mathrm{max}=0.5$ on the residual coordinates and $g_{\mathrm{max}}=1$ on the two map coordinates. Because each residual coordinate is rescaled by one layer and each map coordinate by two, a residual coordinate is stretched by $e^{0.5}$ at most and a map coordinate by at most $e^2$. For $D>64$, the couplings read a fixed random projection $S\in\mathbb{R}^{64\times D}$ of the masked coordinates $u_m$ instead of $u_m$ (the sketch), which preserves invertibility. The two layers that update the map coordinates carry a second network only for the shift (the Fourier branch). This adds the output of an MLP, scaled by $0.25$, to $t_\ell$. The MLP has a single hidden layer of width 256 on the features $\cos(\Omega^\top c+\phi)$ and $\sin(\Omega^\top c+\phi)$ of the conditioner input $c$, with 256 fixed random frequencies $\Omega$ of log-uniform magnitude between 2 and 6, and random phases $\phi$~\cite{Tancik2020}. This branch is initialised to zero and switched on after half of the steps. As it only shifts, it leaves the log-determinant unchanged.

\subsection{Input parameterisation}
\label{sec:input_parametrisation}

To keep raw coordinates from collapsing the training and avoid information loss by applying a PCA to the input, we rotate and rescale all data with
\begin{equation}
z = \Bigl[\tfrac{V^\top (x - \mu) - \mu_h}{\sigma_h},
\tfrac{V_\perp^\top (x - \mu)}{\tilde{\sigma}_t}\Bigr],
\qquad
\tilde{\sigma}_t = \max\bigl(\sigma_t, 10^{-2}\max_j \sigma_{t,j}\bigr),
\label{eq:rawcoords}
\end{equation}
where $V\in\mathbb{R}^{D\times d}$ with $d=\min(50,D-1)$ the number of retained principal components and $V_\perp$ is an orthogonal basis of the remaining components. $\mu$ is the mean of the data, and $\mu_h$ and $\sigma_h$ are the means and standard deviations of the $d$ retained components. $\sigma_t$ are the standard deviations of the remaining components. The whitened block is then rotated so that the layout starts form a plane that is fitted to the order of input distances (Section~\ref{sec:starting_directions}) With $\tilde{\sigma}_t$ we prevent dimensions with small or no variance from blowing up.

\subsection{Starting directions}
\label{sec:starting_directions}
FloDR keeps the global arrangement of the two directions it starts from. On a standardised input, such as the principal components of the paul15 data set and the atlases, every direction has the same variance, which means that the first two principal components are an arbitrary plane that changes with different seeds, leading to seed-dependent changes in the map layout. By default (\texttt{axes="hinge"}), FloDR therefore starts from the plane that keeps the order of input distances the best. We first draw $\min(2{,}000, n/5)$ landmarks by greedy farthest-point sampling in the input, weighting far and rare regions as much as dense ones, and then fit a linear map of the whitened block $Z_h$ to two dimensions to the hinge of equation~\ref{eq:ordinal} on quadruples of landmarks, with Adam at a learning rate of $10^-2$ for 400 steps of $100{,}000$ quadruples each, once from the plane of the first two components and once from each of seven random planes. Frome these eight planes, we keep the one where the orthonormal projection has the lowest hinge on a fixed sample of $200{,}000$ quadruples. We then take $R$ as the orthogonal factor of the QR decomposition of $[Q,I]$, where $Q$ is an orthonormal basis of the plane, and replace the block by $Z_hR$. The fit uses the input metric of the ordinal term and a fixed seed, so that the start depends only on the data. For the data sets we report, this step took 3 to 15~s (on the GPU). Optionally (the heat start, \texttt{axes="heat"}), to start the layout from the heat-diffusion geometry of the neighbour graph, we rotate the whitened principal component block of equation \ref{eq:rawcoords} before training. We compute the heat-geodesic coordinates of HeatGeo on the $k=15$ neighbour lists of the fit~\citep{Huguet2023heat}. We used the settings of the authors, namely, the alpha-decay kernel of PHATE with the distance to the fifth neighbour as bandwidth and a decay of 40, symmetrised, the normalised graph Laplacian $\Delta$, the heat kernel $H=e^{-\tau\Delta}$ at $\tau=10$ approximated by Chebyshev polynomials of order 32 on 500 landmark columns, distances $d_{ij}^{\mathrm{heat}}=\sqrt{-4\tau\log(H_{ij}+10^{-16})}$ and classical multidimensional scaling of the landmarks, on which every data point is then placed. Unlike HeatGeo, we chose the landmarks via greedy farthest-point sampling in the largest connected component to cover rare branches. We regress the first two centred heat coordinates $P$ on the whitened block $Z_h$ by least squares, $B=\arg\min_B||Z_hB-P||$, take $R$ as the orthogonal factor of the QR decomposition of $[B,I]$, and then replace the block by $Z_hR$. Because $R$ is orthogonal in either start, the map stays an exact bijection and the density, the objective, the neighbour graph, and the ordinal target remain unchanged.

\subsection{Training objective}
\label{sec:training_objective}

The objective is a composite of a local and global part and is only applied to the plotted coordinates $y$. First, the local part builds the fuzzy $k$-nearest-neighbour graph of UMAP, with $k=15$ and membership weights $w_{ij}\in(0,1]$, on the measured distances. With $d^Y_{ij} = \lVert y_i - y_j \rVert$ as the embedding distance, a kernel $q_{ij} = \bigl(1 + (d^Y_{ij})^{2b}\bigr)^{-1}$ with $b=0.95$ transforms map distances into affinities. Each edge $(i,j)\in E$ is set from both ends against negatives drawn uniformly for each end in a contrastive (InfoNCE) loss,
\begin{equation}
\begin{gathered}
\mathcal{L}_{\mathrm{local}} = \frac{\sum_{(i,j) \in E} w_{ij}\,\ell_{ij}}{\sum w_{ij}},\\
\ell_{ij} = -\log q_{ij} + \tfrac{1}{2}\Bigl[\log\Bigl(q_{ij} + \sum_{k \in N_i} q_{ik}\Bigr) + \log\Bigl(q_{ij} + \sum_{k \in N_j} q_{jk}\Bigr)\Bigr],
\label{eq:localterm}
\end{gathered}
\end{equation}
with 48 negatives $N_i$ for each end. Unlike repulsion in UMAP, which pushes every sampled pair apart, the push of a negative fades once it is far from the anchor relative to the neighbours of the anchor, leading to gaps closing where the graph links their two sides. the tail exponent $b$ sets the balance, where a heavier tail (smaller $b$) splits points into finer clusters and a lighter tail joins more.

Next, global structure is trained through an ordinal term using only the sign of measured distance differences, as in non-metric MDS~\cite{Kruskal1964MultidimensionalSB}. For two random pairs $(i,j)$ and $(k,l)$, $d^H$ denotes the distance in the input space and $d^Y$ the distance in embedding space. The loss is then the hinge
\begin{equation}
\mathcal{L}_{\mathrm{ord}} =
\mathbb{E}\Bigl[\max\Bigl(0,\operatorname{sgn}\bigl(\log d^H_{kl} - \log d^H_{ij}\bigr)
\bigl(\log d^Y_{ij} - \log d^Y_{kl}\bigr) + \gamma \Bigr)\Bigr],
\label{eq:ordinal}
\end{equation}
with margin $\gamma = 0.1$. Each quadruple $(i,j,k,l)$ only contributes when both pairs have index sums that are not divisible by five, holding out 20\% of pairs for global scoring metrics. As only the sign of the comparison enters, $d^H$ can be any dissimilarity. For binary data, we use the Jaccard distance, detected automatically. Finally, the combined objective is
\begin{equation}
\mathcal{L}=\mathcal{L}_{\mathrm{local}} + \rho(t)(w\mathcal{L}_\mathrm{ord}+w_\mathrm{nll}\mathcal{L}_\mathrm{nll}),
\label{eq:objective}
\end{equation}

where the ramp $\rho(t)$ grows linearly from 0 to 1 over the first $5\%$ of the steps, training local structure first and letting the global contributions fade in. The ordinal weight $w$ is the main user-adjustable parameter of FloDR and trades local neighbourhood preservation against global ordering. The last term $\mathcal{L}_\mathrm{nll}$ is the negative log likelihood of the input under the flow with a diagonal Gaussian base of learned variances, divided by $D$. The released configuration sets $w_\mathrm{nll}=0$ and fits the density after training (see Density). We train with Muon as optimiser~\cite{jordan2024muon} under a single-cycle schedule peaking at three times the base learning rate. Gradients are clipped to norm $5$. The number of steps, the learning rate, and the edge batches are as released for every FloDR fit reported, unless stated otherwise. Unless stated otherwise, every FloDR fit uses $6{,}400$ steps at a base learning rate of $0.0025$, and 48 negative points are sampled for each end of an edge, and four quadruples per point for $\mathcal{L}_\mathrm{ord}$.

\subsection{Local density term}
\label{sec:density_term}
We introduce an optional term from densMAP to keep the local density of the input~\cite{Narayan2021}. The local radius of a point $i$ is the mean-squared length of its edges in the neighbour graph, weighted by their memberships, $R_i=\sum_j w_{ij}d_{ij}^2/\sum_j w_{ij}$, and is taken once with the input distances and once with the map distances. The term is the negative Pearson correlation of $\log R_i$ between the input and the map over all points. It enters equation~(\ref{eq:objective}) with weight $w_\mathrm{dens}$ together with the ramped global terms. It is off by default ($w_\mathrm{dens}=0$). When training on edge batches, the map radii use the length of each edge from the batch that contained it last.

\subsection{Density}

As $f_\theta$ is invertible, the input has an exact density in closed form under the model, that of the output corrected by how much the map stretches volume, $\log p_\theta(x) = \log p_{\mathrm{base}}(f_\theta(x)) + \log \lvert \det J_{f_\theta}(x) \rvert$. We then split the base $p_{\mathrm{base}}(y,r)$ into the two parts $p(y)$ and $p(r \mid y)$, the first a two-dimensional density over the embedding and the second the distribution of the residual $r$ at each position $y$. Drawing a residual from $p(r \mid y)$ and running it backward through the trained map, we get a possible input that would be mapped to $y$. With the map frozen after training, the two parts $p(y)$ and $p(r \mid y)$ are fitted to $(y,r)=f_\theta(x)$, as training the density together with the map overfits the training cells. With $\mathcal{L}_\mathrm{nll}$ added to training ($w_\mathrm{nll}=0.5$, the +NLL arm of the ablations), the training cells of paul15, $80$~\% of the data, collapsed onto a thin sheet in 24 of the 48 residual dimensions, spreading less than a tenth as much as held-out cells around their nearest training cells on the map. The density then rated training cells on average $28$ to $29$ nats above held-out cells, or $11$ to $12$ with the input noise described below, against at most $3$ to $4$ nats without the term (three seeds), while recall on the four benchmark data sets changed by at most $0.004$ (Extended Data Table~\ref{tab:ablations} and Extended Data Fig.~\ref{fig:density}b,c). For $p(y)$, we use a diagonal Gaussian mixture whose number of components is selected from $\{16,32,64,128,256\}$ by held-out likelihood. As affine couplings cannot change the tail class of a Gaussian base~\cite{jaini2020}, we use for $p(r \mid y)$ a Gaussian scale mixture $r=m(y)+s\,\sigma(y)\odot\varepsilon$, with $\varepsilon$ standard normal, $s$ from a logarithmic grid of 20 values between $0.2$ and $3$, and the mixture weights, mean, and log variance read from $y$ by an MLP, standardised by the training map. The calibration of a random half of the training cells, held out from the fit, chooses the tail (this mixture or a diagonal Gaussian), the size of the MLP, and whether to add two coupling layers that reshape $r$ without changing the embedding. The chosen form is then fitted to all training cells. Each held-out cell is ranked among exact draws of the density, by $\log p(y)$ on the map and by $\ell=\log p(x)-\log p(y)$ within the fibre at its position, the set of inputs that the map places there. The ranks are uniform if the density is right. We introduce a coverage certificate that passes when their distance from uniform, the largest difference between the distribution function of the ranks and the diagonal, is below $0.1$ with $95$~\% confidence, refuses when it is above that, and is undecided otherwise. On the bone marrow atlas, fitted on a random 80~\% of the cells, the selected scale mixture kept $0.93$ of the held-out cells inside the $90$~\% regions of their position, whereas a diagonal Gaussian kept $0.87$ and the two added couplings preferred by held-out likelihood kept $0.89$. The certificate refused all three in the fibre and certified the density on the map (Extended Data Fig.~\ref{fig:density}d,e). Both parts are fitted to up to four copies of each cell with isotropic Gaussian input noise of norm half the median distance to the nearest neighbour, its standard deviation along each whitened axis capped at half that of the training cells. Binary inputs get no noise. Held-out cells are scored with one draw of the same noise. The density is then that of the data at a stated resolution and is bounded, so the noise limits how much it can gain by placing training cells on a thin sheet. On the bone marrow, cerebellum, seed and paul15 inputs, the cap binds nowhere, as the noise reaches at most $0.17$, $0.30$, $0.32$ and $0.46$ of an axis's spread. On $10{,}000$ MNIST digits, it binds on all 734 axes outside the 50 leading principal components, and the coverage certificate certified one of three fits and was undecided on the other two. On the bone marrow atlas, the density fitted without noise was refused on all three seeds, as with the noise. The certificate ranks one statistic of each cell and does not always refuse draws that are too wide in the input (see Decoding map positions).

\subsection{Out-of-sample placement}
\label{sec:out_of_sample_placement}

New samples are embedded by a single forward pass $y'=f_\theta(x')_{1:2}$. It keeps the rank correlation of distances but few neighbours, as the layout objective constrains the map only at training points (Extended Data Fig.~\ref{fig:heldout_frontier}). The optional optimised step (\texttt{transform\_opt}), of the same type as the transform functions of UMAP and openTSNE, restores them. Given a new point $x'$, we find its 15 nearest training points in the input metric, initialise its position at the forward pass $y'_0=f_\theta(x')_{1:2}$, and optimise only $y'$ for 200 Adam steps with a learning rate of $0.05$ on the attraction $-\sum_j \log\bigl(1+a\,(d^Y_{y'j})^{2b}\bigr)^{-1}$ to the 15 neighbours, with uniform weights, $(a,b)$ of UMAP at a minimum distance of 0.01 and the training embeddings frozen. Given that a map's scale is arbitrary, the step runs on the map scaled, so that the median distance of the training points to their nearest neighbour is $0.07$. We omit any repulsion. The procedure took about 10~s on the GPU for a random fifth ($20{,}646$ cells) of the bone marrow atlas. The trained map is untouched by this step, and the bijection and the density remain properties of $f_\theta$. We held out 20~\% of the points of each benchmark data set in five random splits to score the placement of unseen points. For atlases with donor annotation, we held out a fifth of the donors in three splits.

\subsection{Global structure beyond the initialisation}
\label{sec:global_structure_beyond}

As every coupling starts as the identity, the initial layout is the fitted plane (see Starting directions). Therefore, the gain in global structure could be inherited from this initial layout rather than produced by the training. We probe this possibility on the paul15 data set, where we replace the fitted plane by a random plane of the whitened block via a random orthogonal rotation, and leave the input, the neighbour graph, and the ordinal target unchanged. Over three seeds, FloDR reaches a CPD over all pairs of $0.722\pm0.002$ against $0.731\pm0.002$ from the fitted start in the same run. This is well above the $0.497$ of the PCA control, which is an arbitrary plane of this input (Extended Data Table~\ref{tab:main-comparison}).

\subsection{Component ablations}
\label{sec:component_ablations}

We ablate the other components of FloDR on the four benchmark data sets with three seeds (Extended Data Table~\ref{tab:ablations}), namely, we ablate the flow depth ($L\in\{2,4,8\}$ coupling layers), the random-projection conditioning of the couplings (with and without sketch), the likelihood ($\mathrm{NLL}$) term of the objective, the Fourier branch, the per-coordinate scale bound (against one scalar gate bounded at 0.5), and the number of steps (800 instead of $6\,400$ at eight times the learning rate, so that the product of learning rate and steps remain the same). All arms stayed within $0.004$ of the base in recall and within $0.007$ in CPD on all four data sets, with the exception of the likelihood term in training on paul15 ($-0.011$ in CPD) and training for 800 steps, which lost $0.003$ to $0.005$ in recall and moved CPD by $-0.019$ to $0.003$. Four coupling layers are therefore enough.

\subsection{Numerical invertibility}
\label{sec:numericalinvertibility}

The round-trip per-point relative $\ell_2$ error $x\mapsto f^{-1}(f(x))$ has a median between $3.0\times10^{-7}$ and $5.7\times10^{-7}$ across the four benchmark data sets and three seeds each, with the worst-case point at $3.4\times10^{-6}$, so the map inverts with single-precision accuracy. In double precision the error stays below $10^{-14}$.

\subsection{Decoding map positions}
\label{sec:decoding}
Because a position does not fix the residual and the inverse amplifies residuals that no training point has (Supplementary Note 1), decoding starts from the mean $z$ of the whitened coordinates (see equation \ref{eq:rawcoords}) of the 15 nearest training points to the position $y$ on a map. Next, it takes Gauss--Newton steps on the two map coordinates of $f_\theta(z)$, the map of the whitened coordinates,
\begin{equation}
z \leftarrow z-\lambda J^{+}\bigl(f_\theta(z)_{1:2}-y\bigr),
\qquad J=\frac{\partial f_\theta(z)_{1:2}}{\partial z},
\label{eq:decode}
\end{equation}
where $ J^{+}$ is the pseudo-inverse of the Jacobian, so that each step is the smallest change of $z$ that closes the linearised gap, and $\lambda=1$ is halved up to six times while the step would widen the gap. Steps stop when the gap is below $10^{-3}$ of the median distance between neighbouring training points on the map, or after a maximum of 50 steps. Finally, the map coordinates of $f_\theta(z)$ are set to $y$ and mapped back so that the map places the returned input exactly at $y$. In our experiments, every position met the tolerance, and the steps raised the mean's error by at most $0.037$ in the units of Extended Data Fig.~\ref{fig:decoding}b.

We then scored the decoding on the splits used in Extended Data Fig.~\ref{fig:heldout_frontier} with $5{,}000$ held-out points per split of an atlas, and without Schneider50k, because UMAP has no inverse transformer under the Jaccard distance. We placed points as in Extended Data Fig.~\ref{fig:decoding}a. Every method then decoded the positions. Settings and errors on every data set are given in Supplementary Note 1. 

\subsection{Data sets and baselines}
\label{sec:data_sets_and_baselines}

The benchmarks are $5{,}000$ images of each MNIST and Fashion-MNIST, the paul15 myeloid progenitors, and $5{,}000$ reactions of the Schneider reaction data set encoded as $2{,}048$-bit DRFP fingerprints~\cite{lecun1998gradient,xiao2017fashion,paul2015transcriptional,schneider2015development,probst2022reaction}. The atlases are of human fetal bone marrow, developing mouse cerebellum, and germinating \textit{Arabidopsis} seed~\cite{jardine2021blood,sepp2023cellular,lee2025single}. Single-cell counts were normalised to $10^4$ per cell, log-transformed, and reduced to the $2{,}000$ ($1{,}000$ for paul15) most variable genes and scaled. Next, we took the 50 first principal components standardised to unit variance, on which the PCA control is an arbitrary projection. All methods used these inputs. The baselines are PCA, UMAP, openTSNE, TriMap, PaCMAP, SQuadMDS, DREAMS, Parametric UMAP, PHATE, PyMDE, and LocalMAP~\cite{Policar2019openTSNEAM,Moon2017PHATEAD,Agrawal2021MinimumDistortionE,Wang2024DimensionRW}, and in Fig.~\ref{fig:frontier} also against UMATO~\cite{jeon2025umato} and  ParamRepulsor~\cite{Huang2024}. All ran at their published defaults, FloDR at $w=2$, with some exceptions. Methods with a number of neighbours used 15 (for PHATE and TriMap we kept 5 neighbours and 12 inliners). UMAP started from PCA because its spectral start lowers its CPD at atlas scale. SQuadMDS is our own implementation of its quartet loss (Adam, $4{,}000$ steps, initialised from PCA at unit scale). For Schneider50k, FloDR, UMAP, openTSNE, and DREAMS used Jaccard distance while the others used Euclidean. No baseline was tuned per data set. We compare further settings in Supplementary Tables 1 and 2.

Recall at $k$ is the mean shore of the $k$ nearest input neighbours of a point among its $k$ nearest neighbours on the map. On atlases, the recall at $k=15$ is computed on $50{,}000$ random anchor cells and CPD is calculated over the held-out pairs among $400{,}000$ random pairs. We pick those whose indices sum to a multiple of five, which are never sampled by the ordinal term. The co-ranking curve is $R_{NX}(k)=((n-1)Q_{NX}(k)-k)/(n-1-k)$, with $Q_{NX}(k)$ the recall at $k$, and its area weights each $k\le n/2$ by $1/k$~\cite{Lee2015}. All other metrics show in Extended Data Table~\ref{tab:main-comparison} are computed on all points at $k=15$. Trustworthiness and continuity~\cite{Venna2001} use the ranks of each point's neighbours on the map in the input and of its input neighbours in the map. Shepard $r$ is the Pearson correlation of all input and map distances. $SNS$ is Kruskal's stress at the least-squares scale of the map, making it independent of the layout size. Centroid correlation is the Spearman correlation of distances between class means in the input and on the map, with class medoids under Jaccard distance. Times are one fit per method on an otherwise idle computer (12 GB NVIDIA RTX 4070 TI GPU and Intel Core i7-13700K).

\subsection{Scaling}
\label{sec:scaling}

We timed one fit per method and size on random subsamples of $10^3$ to $10^6$ cells of a developing zebrafish atlas ($1{,}220{,}178$ cells in total), prepared in the same manner as the bone marrow atlas. Timing includes method package import but not PyTorch import. UMAP ran on one thread, as required when running with seed. FloDR compiles the training step, which adds about 12~s to a first fit.

\subsection{Fast variant}
\label{sec:fast_variant}

The fast variant (\texttt{fast=True}) trains the flow with two coupling layers, where the first updates on half of the residual and the second the map, for $1{,}600$ instead of $6{,}400$ steps at four times the learning rate. No other changes were introduced. On the bone marrow, cerebellum, and seed atlases, the recall and CPD were $0.168$ and $0.814$, $0.135$ and $0.546$, and $0.072$ and $0.710$ (three seeds). We did not evaluate decoding and density on the fast variant.

\subsection{OpenProblems benchmark}
\label{sec:openproblems_benchmark}
We used version 2 of the dimensionality reduction task of OpenProblems~\cite{Luecken2025} (repository task\_dimensionality\_reduction, commit 360499d), and ran its processing and metric components from the published run of 19 December 2024 (commit 7adb5a8) in its container images with tag build\_main. They compute the distances of an embedding between waypoint cells (all cells up to $50{,}000$) and to the centroids of the cell types, score them on 14 metrics (Extended Data Table~\ref{tab:openproblems-metrics}) against the normalised counts of all genes. Each method receives normalised counts. FloDR reduces them to $\min(500, n-1)$ principal components and is fitted outside the container for GPU access with default settings and as a variant with the density term at $w_\mathrm{dens}=2$. The other methods keep their own inputs, from 50 principal components to the full matrix, and their scores from the published run. Each  metric is scaled per data set between the lowest and the highest of three controls (random features, spectral, and true features), clipped to $[0,1]$, and averaged over metrics and data sets. Missing values count as zero as in the leaderboard. In the published run, 150 of the 770 values of the eleven methods on our five data sets are missing. We leave them out for the last column of Table~\ref{tab:openproblems}. Of the 14 data sets in the published run, five have values of at least two of the three controls for every metric required by the scaling. We scored FloDR on all five, namely, mouse myeloid progenitors~\cite{Olsson2016}, mouse HSPC~\cite{Nestorowa2016}, human pancreas~\cite{Luecken2021}, adult mouse primary visual cortex~\cite{Tasic2016}, and zebrafish embryos~\cite{Wagner2018}, with $660$, $1\,920$, $16\,382$, $14\,249$ and $26\,022$ cells, respectively. On the zebrafish data set, the R-based co-ranking component required more main memory than was available, while on the pancreas data set PCA failed due to the distances that the processing component computed being not exactly symmetric. For both, we ported the code to Python to compute the seven co-ranking metrics. The Python port agrees with the R component on the FloDR runs of the pancreas and cortex data sets, and reproduces the published values of PCA on zebrafish and pancreas. To check our runs of the components, we reran PCA on all five data sets. Of its 14 metrics, 11 differ from the published values by less than $5.2\times10^{-4}$. The spectral distance correlation, where the spectral embedding is drawn without a seed, differs by up to $0.11$, and the adjusted Rand index and the normalised mutual information by up to $0.031$ and $0.014$, due to the Leiden clusters, on which they are computed, differing between the published run and ours. Combined these change the score of PCA by up to $0.009$ per data set. Note that we keep all metrics and data sets nevertheless. Version 1 of the task (commit 65efdc8 in the openproblems repository), which scores four data sets on ten metrics, has its results published on the OpenProblems website. We copied the metric functions, data loaders, and normalisation without changes and ran them in an environment set up according to the requirements of the task (Python 3.8, NumPy 1.23.5, SciPy 1.9.3, scikit-learn 1.1.3, AnnData 0.8.0, scanpy 1.9.1, umap-learn 0.5.3 and pynndescent 0.5.8). FloDR is fitted as in version 2. Scaling the published metric values per data set between the three baselines of the task reproduces the published scaled scores. The two distance correlations that draw random numbers without a seed differed by a maximum of $0.004$ and $0.023$, respectively, when rerunning PCA. The density preservation differed by a maximum of $0.019$. Altogether, these differences changed the score of PCA per data set by $0.0004$ at most.

\subsection{Tests against unseen measurements}
Every test relates map distances to a measurement that no map was given. The fate of the clone of a cell and the coupling of cell types within clones (LARRY~\cite{Weinreb2020}, and the LSK and reprogramming arms of CellTag-multi~\cite{Jindal2023}), protein levels (CITE-seq~\cite{luecken2021sandbox}), distances between sampling sites~\cite{Ramachandran2005} (HGDP-CEPH genotypes~\cite{Bergstrm2020}, seven of 54 populations held out and placed on the map), and cortical depth (SEA-AD MERFISH data~\cite{Gabitto2024}). Maps shared 50 unscaled principal components and the settings above. FloDR was compared with seven others (four in the placement test) over five seeds on LARRY and the genotypes and three elsewhere (Supplementary Note~2).

Clones, populations or donors, and seeds of each map were resampled $2\,000$ times ($1\,000$ for the verdicts of FloDR in the LSK and reprogramming tests). FloDR won against a map when the $2.5$~\% quantile of the paired difference was positive and the mean at least $0.02$. It lost in the mirror case and tied otherwise. For protein, a Wilcoxon test over batches decided ($p<0.05$, median difference $\geq0.02$). A primary readout (fate, coupling on LARRY) and any readout of the genotype and cortex tests supported no claim when the input did not exceed UMAP by $0.03$, as such a test cannot reward a map for keeping the input. The non-metric MDS~\cite{Kruskal1964MultidimensionalSB} is the scikit-learn implementation, started from two principal components.

\subsection{Statistics and reproducibility}
\label{sec:statistics}

Benchmark and atlas scores are means over seeds 0, 1, and 2 with standard deviations where given. Pairwise-distance panels and co-ranking curves always show seed 0. OpenProblems scores come from one fit per data set. Held-out placement and decoding use five random splits of each benchmark data set and three by-donor or by-sample splits of the atlases. The coverage certificate uses the Dvoretzky-Kiefer-Wolfowitz band at $\alpha=0.05$~\cite{Massart1990}.

\subsection{Data availability}
All data sets used are public. MNIST and Fashion-MNIST from OpenML (mnist\_784 and Fashion-MNIST, version 1), Schneider50k from \url{https://github.com/reymond-group/drfp}, paul15 through scanpy, the bone marrow, cerebellum, and seed atlases~\cite{jardine2021blood,sepp2023cellular,lee2025single}, the zebrafish atlas (data set bcd4103c-afdc-4233-af6e-554589d70cd3) from CELLxGENE Discover (\url{https://cellxgene.cziscience.com}), and the OpenProblems data and scores from \url{https://openproblems.bio}. The HGDP-CEPH genomes as PLINK 2 files (\url{https://www.cog-genomics.org/plink/2.0/resources}), LARRY from GEO (GSE140802), CellTag-multi from GEO (GSE216606 and GSE216518) with the clone tables (\url{https://github.com/morris-lab/CellTag-multi-2023}) and annotated cells (figshare 26367055, Zenodo 22715159) of its authors, CITE-seq from GEO (GSE194122), and the SEA-AD MERFISH data of 11 December 2024 from \url{https://registry.opendata.aws/allen-sea-ad-atlas/}.

\subsection{Code availability}

FloDR is available as a Python package (\texttt{flodr}) from the Python Package Index and at \url{https://github.com/daenuprobst/flodr} under the MIT licence.

\subsection{Use of large language models}
As non-native English speakers, we used local LLMs for grammar, sentence structure, and other proofreading, plus as an editor to suggest paragraphs to be shortened or moved. Furthermore, we used local LLMs for enhanced autocomplete and documenting the code. Finally, we used the Gemini LLM provided by Google for search.

\paragraph{Acknowledgements.} We would like to thank the OpenProblems team and the authors of all other data sets used here for making their data, code, and results publicly available.

\paragraph{Funding.} We do not have a funder or a grant number to declare.

\paragraph{Author contributions.} D.P. conceived FloDR and implemented the initial architecture. A.B. contributed improvements to the architecture and evaluated the decoding of map positions using single-cell RNA-seq data. Both contributed equally to other tasks.

\paragraph{Competing interestes.} The authors declare no competing interests.





\bibliography{sn-bibliography}% common bib file

\clearpage
\section*{Extended Data}

\setcounter{figure}{0}
\setcounter{table}{0}
\renewcommand{\figurename}{Extended Data Figure}
\renewcommand{\tablename}{Extended Data Table}
\renewcommand{\theHfigure}{ED\arabic{figure}}
\renewcommand{\theHtable}{ED\arabic{table}}

\begin{figure}[h!]
\centering
\includegraphics[width=\textwidth,height=0.86\textheight,keepaspectratio]{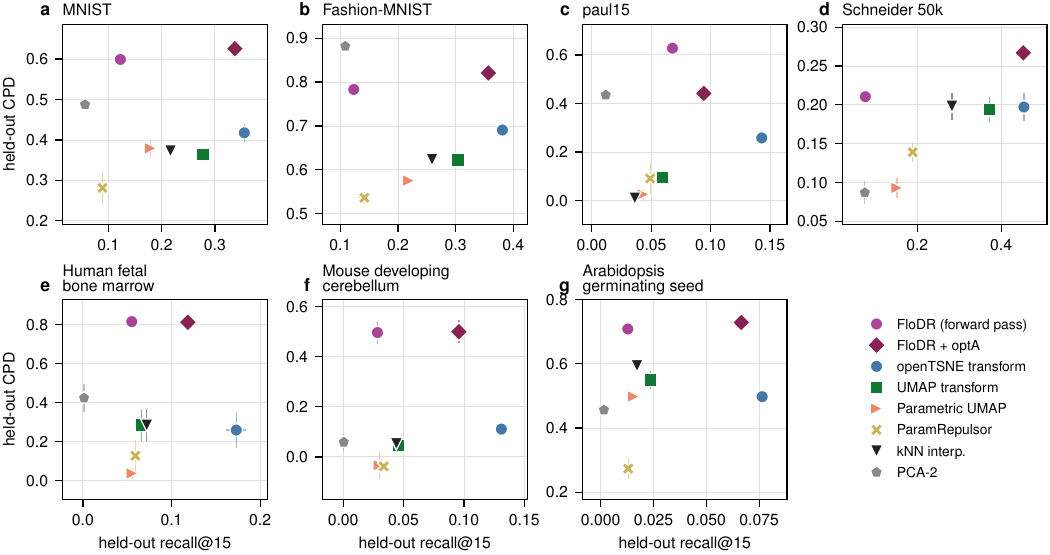}
\caption{\textbf{Placement of held-out points.} Recall at $k=15$ and CPD of the held-out points only, on the four benchmark data sets (\textbf{a}--\textbf{d}, five random splits with 20~\% of the points held out), and the three atlases (\textbf{e}-\textbf{g}, three splits with a fifth of the donors or samples held out). Points are means and error bars standard deviations over the splits. The upper-right corner is better on both axes. The forward pass of FloDR keeps the rank correlation of distances but few neighbours (0.07 to 0.12 on the benchmark data sets). The optimised step (FloDR $+$ optA) raises the recall substantially on all data sets, while keeping CPD above that of every other method except PCA on Fashion-MNIST. On the atlases, it keeps more neighbours than every other method except openTSNE. kNN interpolation is a control that places a point at the inverse-distance weighted mean of the UMAP positions of its 15 nearest training points in the input.}
\label{fig:heldout_frontier}
\end{figure}

\begin{figure}[h!]
\centering
\includegraphics[width=\textwidth,height=0.86\textheight,keepaspectratio]{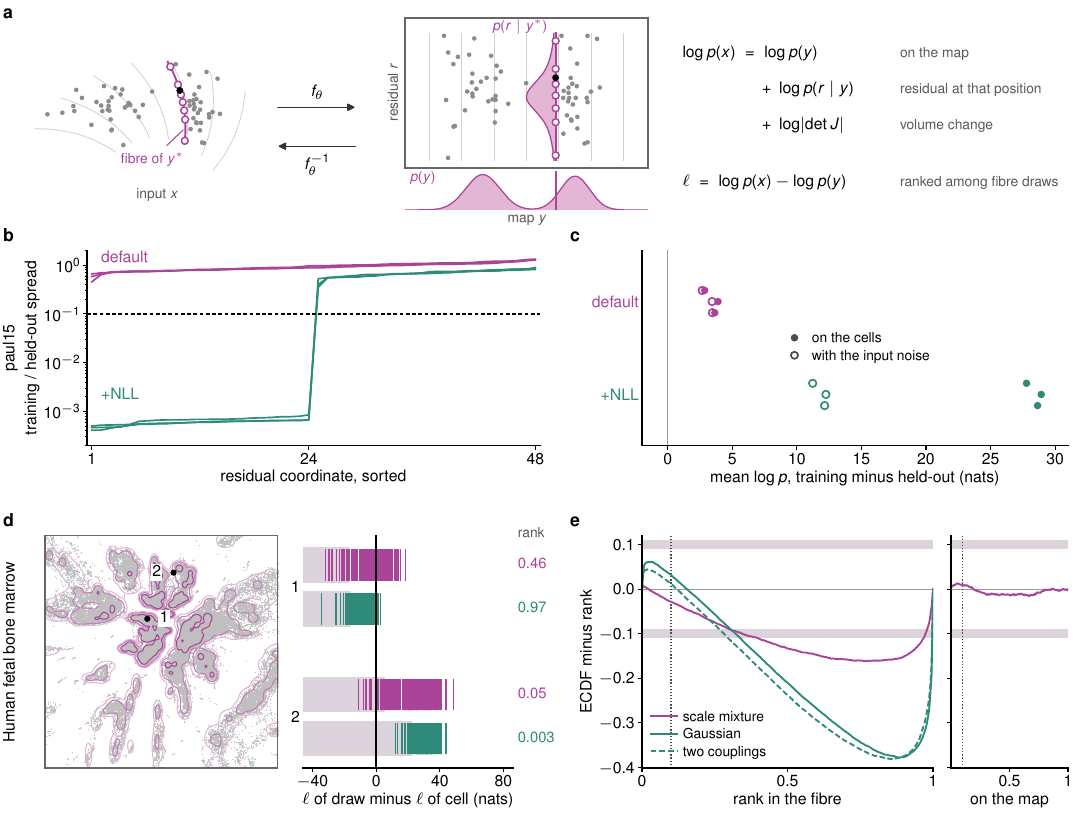}
\caption{\textbf{The density of FloDR and its check on held-out cells.} \textbf{a}, Toy data with a two-dimensional input and a one-dimensional map. Each grey curve is a fibre, the inputs that $f_\theta$ maps to one position $y$, and becomes a vertical line in $(y,r)$. Draws from $p(r\mid y^*)$ (open circles) run back through $f_\theta^{-1}$ to inputs on the fibre of $y^*$, and a held-out cell (black) is ranked among them by $\ell=\log p(x)-\log p(y)$. In FloDR, $y$ has two and $r$, the residual, has $D-2$ coordinates. \textbf{b}, paul15, fitted on $80$~\% of the cells with three seeds. Per residual coordinate, the spread of the training cells around the mean residual of their 15 nearest training cells on the map, divided by that of the held-out cells, with default FloDR settings (magenta) and with $\mathcal{L}_\mathrm{nll}$ during training (+NLL, teal). The dashed line is at $0.1$. \textbf{c}, Mean $\log p$ of training minus held-out cells, on the cells (filled) and with one draw of the input noise (open). \textbf{d}, Two held-out cells of the bone marrow atlas, fitted on $80$~\% of the cells, on its map (training cells in grey, the $50$, $80$ and $95$~\% isolines of $p(y)$ in magenta), and $\ell$ of 199 exact draws from the fibre at each cell minus $\ell$ of the cell, under the selected scale mixture (magenta) and a diagonal Gaussian (teal). Shading marks the lowest tenth of the draws, so a cell in it lies outside its $90$~\% region. The rank is $(b+\tfrac{1}{2})/200$ for $b$ of these draws below the cell. With its own draws, the certificate ranks cell 1 (erythroid) at $0.49$ and $0.98$ and cell 2 (B lineage) at $0.05$ and $0.005$ under the two forms. Cell 1 is nearest $0.5$ under the scale mixture among the cells the Gaussian ranks above $0.97$, and cell 2 is the B-lineage cell nearest $0.05$ among those ranked $0.03$ to $0.07$ by the scale mixture and below $0.005$ by the Gaussian. \textbf{e}, Share of the $20{,}646$ held-out cells ranked below $u$, minus $u$, in the fibre for the three forms of $p(r\mid y)$ and on the map, where $p(y)$ is the same for all three. The bands lie at $\pm0.1$ with the half-width of the $95$~\% confidence band, $0.0095$: a curve inside their inner edges is certified, and one beyond their outer edges refused. The share of cells right of the dotted line is the share inside the $90$~\% regions.}
\label{fig:density}
\end{figure}

\begin{figure}[h!]
\centering
\includegraphics[width=\textwidth,height=0.86\textheight,keepaspectratio]{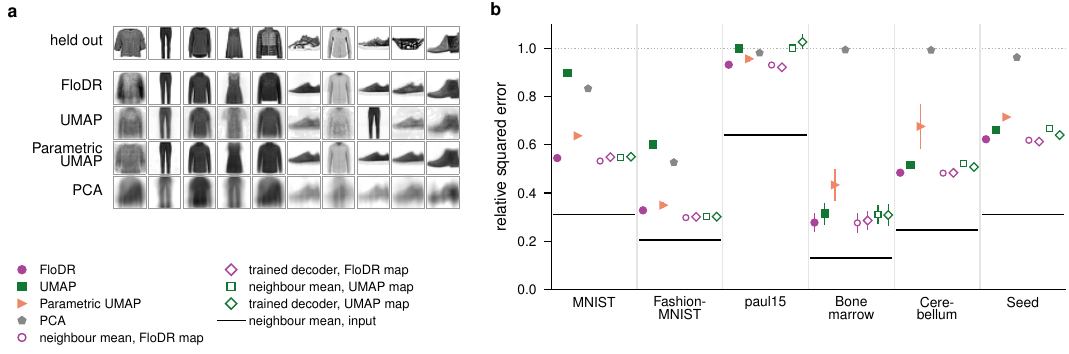}
\caption{\textbf{Decoding positions of the map.} \textbf{a}. The first held-out image of each class from Fashion-MNIST in one split (top row) and what each method returns for the position at which it placed the image (rows 2--5). FloDR places the images by its optimised step and UMAP by its transform, both are using the nearest training images in the input. Parametric UMAP places images by its encoder and PCA by projection, without using neighbours. FloDR returns the mean of the 15 nearest training point on the map, moved onto the fibre of the position, UMAP its inverse transform, which starts from a weighted mean of 784 images around the position, Parametric UMAP its decoder, and PCA the projection back from the first two principal components. The inverse transform of UMAP depends on the number of threads. For the eighth image, it returned trousers with 2 and 8 threads (shown with 8), and a sneaker with 4 and 20 (Supplementary Note 1). \textbf{b}. Squared error of decoded held-out points over that of the mean of the training points. 0 is exact and 1 is no better than the mean. Symbols are means and bars standard deviations over five random splits of MNIST, Fashion-MNIST, and paul15, and three splits of the atlases by donor or sample. Filled symbols are the four decodes of panel a. When placed only by its forward pass, using no neighbours, FloDR decoded less accurately than Parametric UMAP on 19 of 24 splits (not drawn, Supplementary Note 1). Open symbols show what a layout allows, on the maps of FloDR and UMAP, namely, the mean of the 15 training points nearest to the position, and a decoder trained on the layout. The black horizontal lines are the mean of the 15 nearest training points in the input, which does not use a map.}
\label{fig:decoding}
\end{figure}

\begin{figure}[h!]
\centering
\includegraphics[width=0.97\textwidth,height=0.86\textheight,keepaspectratio]{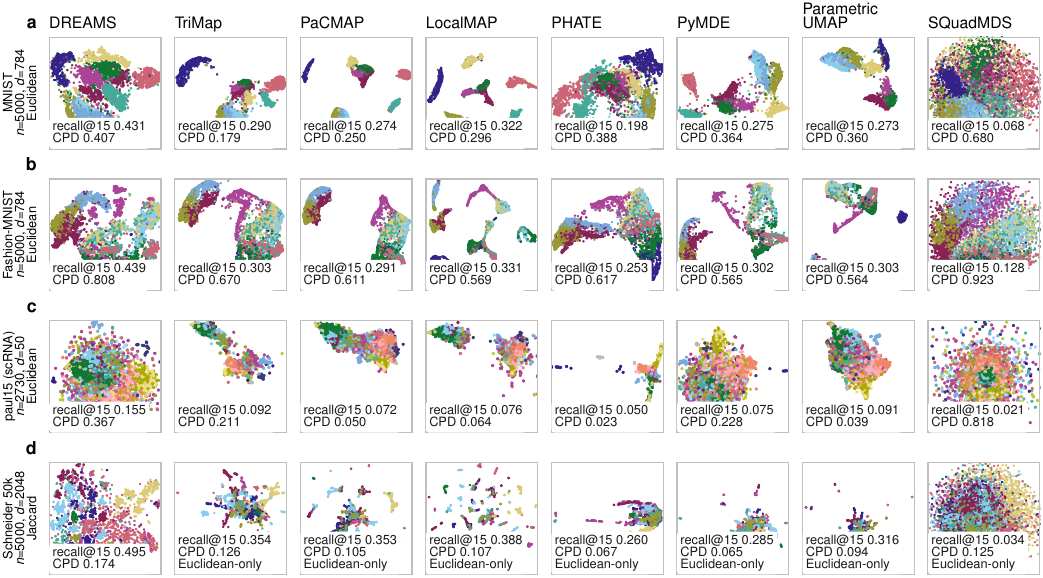}
\caption{\textbf{Remaining baselines on the four benchmark data sets.} SQuadMDS attains the highest CPD of any evaluated method on three of four data sets, while keeping fewer neighbours than every method other than PCA, and fewer than PCA on Schneider50k. This trade-off is summarised in Fig.~\ref{fig:frontier}.}
\label{fig:comparison_appendix}
\end{figure}

\begin{figure}[h!]
\centering
\includegraphics[width=\textwidth,height=0.86\textheight,keepaspectratio]{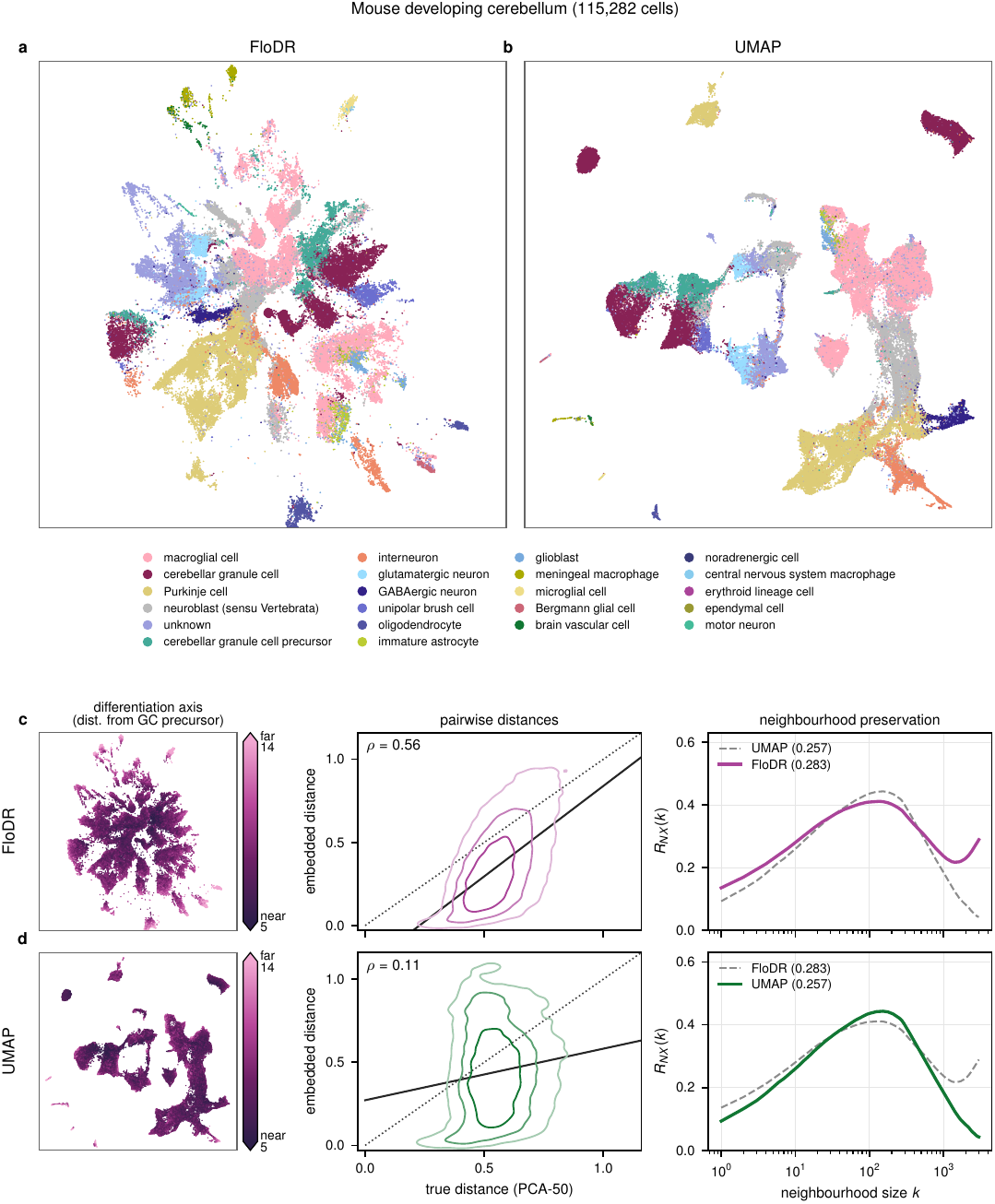}
\caption{\textbf{FloDR and UMAP embeddings of the developing mouse cerebellum atlas} ($n = 115{,}282$ cells, 22 annotated cell types). Panels are arranged as in Fig.~\ref{fig:bmarrow}. The Spearman correlation of $60\,000$ randomly sampled pairs shown is $0.56$ for FloDR versus $0.11$ for UMAP. The log-weighted areas under the co-ranking curves are $0.283$ and $0.257$, respectively. The maps of FloDR keep the order of the median distances between the annotated types (Spearman correlation $0.76$ to $0.80$ over three fits, against $0.32$ to $0.38$ for UMAP, on the 20 types with at least 50 cells) and reproduce the distance from the granule cell precursors ($0.53$ to $0.55$, against $-0.09$ to $0.08$, panels \textbf{c}, \textbf{d}). Like in the input, FloDR places granule cells and neuroblasts at the centre and the precursors beside the granule cells, with $92$ to $93$\% of the granule lineage in one piece, against $56$ to $57$\% for UMAP. UMAP places precursors and granule cells far apart and splits off three pieces of a single sample on every fit (adult granule cells, and the macroglia and neuroblasts of one of the three earliest samples), FloDR at most two. FloDR keeps the Purkinje cells in one piece and orders Purkinje cells and macroglia by developmental age on every fit. It leaves interneurons and macroglia less contiguous than UMAP and sets glutamatergic neurons farther from unipolar brush cells than the input.}

\label{fig:cerebellum}
\end{figure}

\begin{figure}[h!]
\centering
\includegraphics[width=\textwidth,height=0.86\textheight,keepaspectratio]{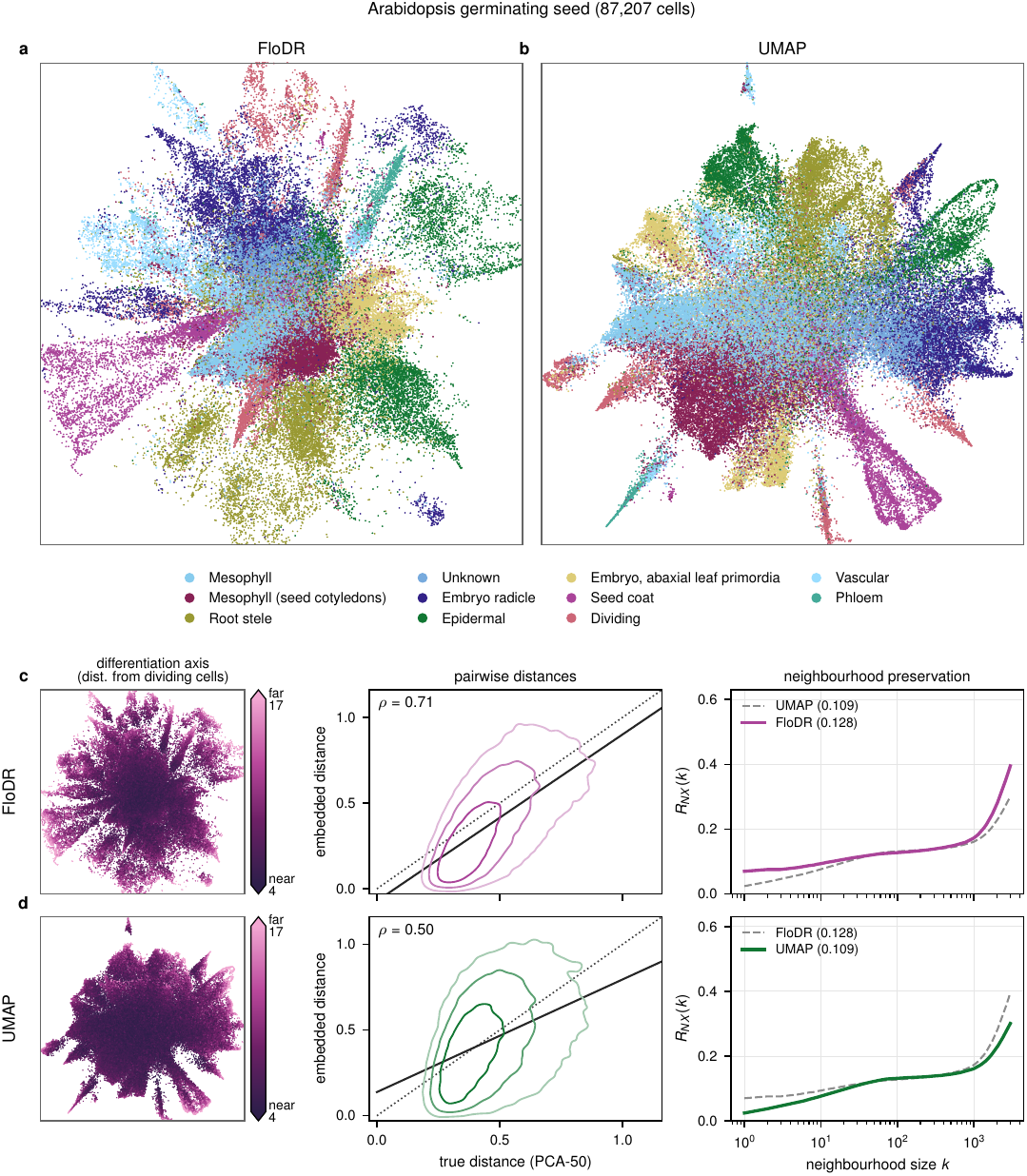}
\caption{\textbf{FloDR and UMAP embeddings of the germinating \textit{Arabidopsis} seed atlas} ($n = 87{,}207$ cells, 11 annotated cell types). Panels are arranged as in Fig.~\ref{fig:bmarrow}. The Spearman correlation of the $59{,}999$ randomly sampled pairs is $0.71$ for FloDR versus $0.50$ for UMAP. The log-weighted areas under the co-ranking curves are $0.128$ and $0.109$, respectively. The maps of FloDR keep the order of the distances between the ten annotated types other than the unknown cells (Spearman correlation $0.77$ to $0.79$ over three fits, against $0.45$ to $0.65$ for UMAP), also among the most similar third of the pairs ($0.69$ to $0.82$ against $0.51$ to $0.70$). Like in the input, the two mesophyll types and the abaxial leaf primordia form the centre of every map of FloDR, with $69$ to $70$\% of the primordia in one piece, compared to $29$ to $33$\% for UMAP. UMAP puts vascular or seed-coat cells at its centre instead of the primordia and moves the seed coat away from the mesophyll. Both maps show the distance from the dividing cells (a cell-cycle state spread over several tissues) equally. FloDR splits the embryo radicle into more pieces.}

\label{fig:plant}
\end{figure}

\begin{figure}[h!]
\centering
\includegraphics[width=\textwidth,keepaspectratio]{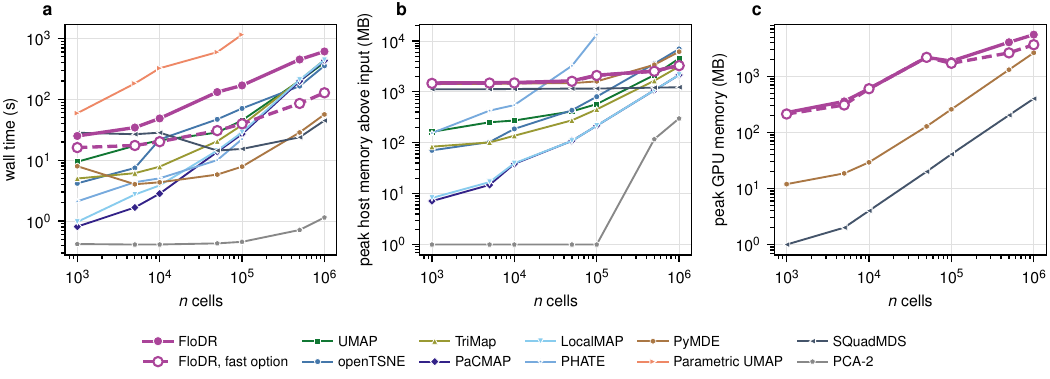}
\caption{\textbf{Wall time and peak memory against the number of cells} \textbf{a}. Wall time. \textbf{b}. Peak main memory above that of the input. \textbf{c}. Peak GPU memory of FloDR, PyMDE, and SQuadMDS. Single fit per method and size on random subsamples of the 50 principal components of the zebrafish atlas, under conditions described in Methods. FloDR runs at is default and fast variants. Lines end where a method failed. PHATE ran out of main memory and Parametric UMAP out of GPU memory.}
\label{fig:scaling}
\end{figure}

\begin{table}[h!]
\caption{\textbf{FloDR against the baselines on the four benchmark data sets.} Recall at $k=15$, trustworthiness (Trust), continuity (Cont.), CPD on the held-out pairs, Shepard $r$, scale-normalised stress (SNS, lower is better) and the centroid correlation (Centr., see Methods). Means over three seeds, with the standard deviation for recall, CPD and SNS. The best value of a score column within a data set is in bold. Time is that of one fit in seconds, with the training step of FloDR compiled (Methods), and DREAMS was not timed. FloDR, fast, is the fast variant. $\dagger$, fitted on the Euclidean distance of the fingerprints and scored on the Jaccard distance.}
\label{tab:main-comparison}
{\footnotesize\setlength{\tabcolsep}{1.6pt}\chg{\renewcommand{\arraystretch}{0.82}}
\begin{tabular}{@{}lcccccccr@{}}
\toprule
Method & Recall & Trust & Cont. & CPD & Shepard $r$ & SNS & Centr. & Time (s) \\
\midrule
\multicolumn{9}{@{}l}{\textit{MNIST ($n=5\,000$, $d=784$, pixels, Euclidean)}} \\
FloDR, $w=2$ & $0.431 \pm 0.000$ & $0.974$ & $0.952$ & $0.636 \pm 0.000$ & $0.631$ & $0.415 \pm 0.001$ & $0.564$ & $53.5$ \\
FloDR, fast & $0.430 \pm 0.000$ & $0.974$ & $0.952$ & $0.640 \pm 0.001$ & $0.632$ & $0.418 \pm 0.001$ & $0.535$ & $20.9$ \\
FloDR, $w=0$ & $\mathbf{0.435 \pm 0.001}$ & $0.973$ & $0.955$ & $0.477 \pm 0.009$ & $0.496$ & $0.432 \pm 0.011$ & $0.514$ & $53.1$ \\
UMAP & $0.339 \pm 0.001$ & $0.955$ & $\mathbf{0.962}$ & $0.363 \pm 0.008$ & $0.401$ & $0.440 \pm 0.006$ & $0.543$ & $15.5$ \\
openTSNE & $\mathbf{0.435 \pm 0.000}$ & $\mathbf{0.975}$ & $0.959$ & $0.415 \pm 0.002$ & $0.445$ & $0.393 \pm 0.000$ & $0.729$ & $7.5$ \\
DREAMS & $0.431 \pm 0.000$ & $0.971$ & $0.956$ & $0.407 \pm 0.001$ & $0.440$ & $0.406 \pm 0.000$ & $0.796$ & -- \\
PCA & $0.057 \pm 0.000$ & $0.733$ & $0.917$ & $0.489 \pm 0.000$ & $0.523$ & $0.413 \pm 0.000$ & $\mathbf{0.797}$ & $0.1$ \\
TriMap & $0.290 \pm 0.002$ & $0.937$ & $0.949$ & $0.179 \pm 0.005$ & $0.188$ & $0.523 \pm 0.001$ & $0.730$ & $6.1$ \\
PaCMAP & $0.274 \pm 0.002$ & $0.938$ & $0.952$ & $0.250 \pm 0.008$ & $0.295$ & $0.463 \pm 0.003$ & $0.692$ & $2.2$ \\
LocalMAP & $0.322 \pm 0.001$ & $0.960$ & $0.949$ & $0.296 \pm 0.007$ & $0.340$ & $0.428 \pm 0.001$ & $0.748$ & $2.1$ \\
PHATE & $0.198 \pm 0.002$ & $0.856$ & $0.959$ & $0.388 \pm 0.005$ & $0.422$ & $0.437 \pm 0.007$ & $0.687$ & $7.3$ \\
PyMDE & $0.275 \pm 0.003$ & $0.936$ & $0.961$ & $0.364 \pm 0.004$ & $0.400$ & $0.428 \pm 0.004$ & $0.587$ & $4.6$ \\
Param.\ UMAP & $0.273 \pm 0.002$ & $0.943$ & $\mathbf{0.962}$ & $0.360 \pm 0.007$ & $0.394$ & $0.457 \pm 0.009$ & $0.548$ & $454.1$ \\
SQuadMDS & $0.068 \pm 0.000$ & $0.781$ & $0.906$ & $\mathbf{0.680 \pm 0.001}$ & $\mathbf{0.679}$ & $\mathbf{0.365 \pm 0.000}$ & $0.713$ & $25.8$ \\
\midrule
\multicolumn{9}{@{}l}{\textit{Fashion-MNIST ($n=5\,000$, $d=784$, pixels, Euclidean)}} \\
FloDR, $w=2$ & $0.434 \pm 0.001$ & $\mathbf{0.985}$ & $0.980$ & $0.823 \pm 0.001$ & $0.815$ & $0.301 \pm 0.001$ & $0.888$ & $54.0$ \\
FloDR, fast & $0.435 \pm 0.001$ & $\mathbf{0.985}$ & $0.980$ & $0.822 \pm 0.001$ & $0.818$ & $0.297 \pm 0.001$ & $0.893$ & $21.1$ \\
FloDR, $w=0$ & $0.435 \pm 0.002$ & $0.984$ & $0.982$ & $0.596 \pm 0.003$ & $0.583$ & $0.428 \pm 0.002$ & $0.890$ & $53.6$ \\
UMAP & $0.350 \pm 0.002$ & $0.979$ & $0.983$ & $0.604 \pm 0.007$ & $0.605$ & $0.395 \pm 0.005$ & $0.917$ & $15.2$ \\
openTSNE & $\mathbf{0.442 \pm 0.001}$ & $\mathbf{0.985}$ & $0.982$ & $0.699 \pm 0.014$ & $0.707$ & $0.320 \pm 0.003$ & $0.913$ & $7.1$ \\
DREAMS & $0.439 \pm 0.001$ & $0.984$ & $0.978$ & $0.808 \pm 0.001$ & $0.810$ & $0.289 \pm 0.000$ & $\mathbf{0.977}$ & -- \\
PCA & $0.107 \pm 0.000$ & $0.915$ & $0.972$ & $0.880 \pm 0.000$ & $0.882$ & $0.267 \pm 0.000$ & $0.974$ & $0.1$ \\
TriMap & $0.303 \pm 0.002$ & $0.971$ & $\mathbf{0.984}$ & $0.670 \pm 0.002$ & $0.667$ & $0.369 \pm 0.001$ & $0.955$ & $6.4$ \\
PaCMAP & $0.291 \pm 0.001$ & $0.973$ & $0.981$ & $0.611 \pm 0.003$ & $0.592$ & $0.425 \pm 0.001$ & $0.912$ & $2.0$ \\
LocalMAP & $0.331 \pm 0.004$ & $0.979$ & $0.971$ & $0.569 \pm 0.014$ & $0.580$ & $0.367 \pm 0.004$ & $0.838$ & $2.0$ \\
PHATE & $0.253 \pm 0.002$ & $0.951$ & $0.983$ & $0.617 \pm 0.005$ & $0.607$ & $0.410 \pm 0.008$ & $0.931$ & $8.1$ \\
PyMDE & $0.302 \pm 0.001$ & $0.974$ & $0.982$ & $0.565 \pm 0.096$ & $0.576$ & $0.395 \pm 0.035$ & $0.850$ & $4.7$ \\
Param.\ UMAP & $0.303 \pm 0.002$ & $0.975$ & $0.982$ & $0.564 \pm 0.003$ & $0.550$ & $0.444 \pm 0.001$ & $0.918$ & $400.2$ \\
SQuadMDS & $0.128 \pm 0.001$ & $0.930$ & $0.973$ & $\mathbf{0.923 \pm 0.000}$ & $\mathbf{0.916}$ & $\mathbf{0.251 \pm 0.000}$ & $0.970$ & $26.7$ \\
\midrule
\multicolumn{9}{@{}l}{\textit{paul15 ($n=2\,730$, $d=50$, standardised principal components, Euclidean)}} \\
FloDR, $w=2$ & $0.168 \pm 0.001$ & $\mathbf{0.715}$ & $0.815$ & $0.730 \pm 0.002$ & $0.747$ & $0.364 \pm 0.000$ & $\mathbf{0.843}$ & $30.1$ \\
FloDR, fast & $0.165 \pm 0.001$ & $0.713$ & $0.815$ & $0.729 \pm 0.001$ & $0.747$ & $0.361 \pm 0.000$ & $0.835$ & $17.5$ \\
FloDR, $w=0$ & $\mathbf{0.176 \pm 0.001}$ & $\mathbf{0.715}$ & $0.802$ & $0.269 \pm 0.014$ & $0.279$ & $0.483 \pm 0.004$ & $0.591$ & $29.5$ \\
UMAP & $0.102 \pm 0.000$ & $0.656$ & $0.789$ & $0.049 \pm 0.003$ & $0.043$ & $0.548 \pm 0.000$ & $0.341$ & $11.7$ \\
openTSNE & $0.169 \pm 0.001$ & $0.709$ & $0.795$ & $0.327 \pm 0.006$ & $0.319$ & $0.463 \pm 0.001$ & $0.512$ & $5.3$ \\
DREAMS & $0.155 \pm 0.001$ & $0.711$ & $0.802$ & $0.367 \pm 0.015$ & $0.377$ & $0.446 \pm 0.004$ & $0.650$ & -- \\
PCA & $0.015 \pm 0.001$ & $0.550$ & $0.726$ & $0.497 \pm 0.038$ & $0.568$ & $0.451 \pm 0.007$ & $0.744$ & $0.0$ \\
TriMap & $0.092 \pm 0.000$ & $0.692$ & $0.721$ & $0.211 \pm 0.013$ & $0.203$ & $0.590 \pm 0.002$ & $0.380$ & $5.0$ \\
PaCMAP & $0.072 \pm 0.000$ & $0.664$ & $0.743$ & $0.050 \pm 0.003$ & $0.048$ & $0.615 \pm 0.002$ & $0.309$ & $0.8$ \\
LocalMAP & $0.076 \pm 0.001$ & $0.679$ & $0.739$ & $0.064 \pm 0.005$ & $0.044$ & $0.643 \pm 0.003$ & $0.290$ & $1.4$ \\
PHATE & $0.050 \pm 0.000$ & $0.603$ & $0.754$ & $0.023 \pm 0.001$ & $0.161$ & $0.840 \pm 0.001$ & $0.479$ & $5.1$ \\
PyMDE & $0.075 \pm 0.002$ & $0.631$ & $\mathbf{0.818}$ & $0.228 \pm 0.007$ & $0.217$ & $0.509 \pm 0.003$ & $0.453$ & $4.6$ \\
Param.\ UMAP & $0.091 \pm 0.001$ & $0.652$ & $0.787$ & $0.039 \pm 0.002$ & $0.053$ & $0.556 \pm 0.006$ & $0.332$ & $100.6$ \\
SQuadMDS & $0.021 \pm 0.000$ & $0.577$ & $0.761$ & $\mathbf{0.818 \pm 0.000}$ & $\mathbf{0.853}$ & $\mathbf{0.312 \pm 0.000}$ & $0.792$ & $26.5$ \\
\midrule
\multicolumn{9}{@{}l}{\textit{Schneider50k ($n=5\,000$, $d=2\,048$, reaction fingerprints, Jaccard)}} \\
FloDR, $w=2$ & $0.516 \pm 0.000$ & $0.968$ & $0.933$ & $\mathbf{0.287 \pm 0.004}$ & $\mathbf{0.320}$ & $0.430 \pm 0.000$ & $0.210$ & $119.1$ \\
FloDR, fast & $0.514 \pm 0.002$ & $0.967$ & $0.933$ & $0.275 \pm 0.004$ & $0.308$ & $0.428 \pm 0.000$ & $0.215$ & $54.3$ \\
FloDR, $w=0$ & $\mathbf{0.518 \pm 0.001}$ & $0.967$ & $0.936$ & $0.184 \pm 0.006$ & $0.244$ & $0.450 \pm 0.001$ & $0.137$ & $118.7$ \\
UMAP & $0.437 \pm 0.001$ & $0.957$ & $\mathbf{0.943}$ & $0.189 \pm 0.014$ & $0.246$ & $0.461 \pm 0.007$ & $0.426$ & $16.0$ \\
openTSNE & $0.499 \pm 0.001$ & $0.960$ & $0.940$ & $0.199 \pm 0.004$ & $0.274$ & $\mathbf{0.426 \pm 0.000}$ & $0.436$ & $20.0$ \\
DREAMS & $0.495 \pm 0.000$ & $0.957$ & $0.922$ & $0.174 \pm 0.004$ & $0.235$ & $0.440 \pm 0.001$ & $0.261$ & -- \\
PCA$^{\dagger}$ & $0.075 \pm 0.000$ & $0.734$ & $0.796$ & $0.083 \pm 0.000$ & $0.125$ & $0.548 \pm 0.000$ & $0.288$ & $0.6$ \\
TriMap$^{\dagger}$ & $0.354 \pm 0.001$ & $0.933$ & $0.873$ & $0.126 \pm 0.011$ & $0.071$ & $0.786 \pm 0.024$ & $0.408$ & $6.7$ \\
PaCMAP$^{\dagger}$ & $0.353 \pm 0.001$ & $0.946$ & $0.886$ & $0.105 \pm 0.002$ & $0.178$ & $0.482 \pm 0.001$ & $0.372$ & $3.3$ \\
LocalMAP$^{\dagger}$ & $0.388 \pm 0.003$ & $\mathbf{0.973}$ & $0.884$ & $0.107 \pm 0.002$ & $0.179$ & $0.458 \pm 0.001$ & $0.284$ & $3.4$ \\
PHATE$^{\dagger}$ & $0.260 \pm 0.001$ & $0.812$ & $0.857$ & $0.067 \pm 0.003$ & $0.075$ & $0.747 \pm 0.008$ & $0.419$ & $7.6$ \\
PyMDE$^{\dagger}$ & $0.285 \pm 0.001$ & $0.869$ & $0.873$ & $0.065 \pm 0.003$ & $0.128$ & $0.559 \pm 0.001$ & $0.241$ & $5.1$ \\
Param.\ UMAP$^{\dagger}$ & $0.316 \pm 0.005$ & $0.911$ & $0.891$ & $0.094 \pm 0.013$ & $0.139$ & $0.598 \pm 0.008$ & $\mathbf{0.459}$ & $296.8$ \\
SQuadMDS$^{\dagger}$ & $0.034 \pm 0.000$ & $0.692$ & $0.784$ & $0.125 \pm 0.000$ & $0.178$ & $0.445 \pm 0.000$ & $0.438$ & $26.7$ \\
\botrule
\end{tabular}}
\end{table}

\begin{table}[h!]
\caption{\textbf{Component ablations.} Neighbourhood recall at $k=15$ and CPD on the held-out pairs. Mean $\pm$ standard deviation over three seeds, on the four benchmark data sets of Fig.~\ref{fig:comparison}. Base is FloDR at its default ($w=2$), and every other arm changes one component of it. Two or eight instead of four coupling layers, no random-projection sketch in the couplings, the likelihood term in training (+NLL, $w_\mathrm{nll}=0.5$), no Fourier branch, one scalar gate instead of a scale bound per coordinate, and 800 instead of $6\,400$ steps. The sketch only acts when inputs have more than 64 dimensions, so the no-sketch arm is not run on the 50-dimensional paul15 input.}
\label{tab:ablations}
{\setlength{\tabcolsep}{6pt}
\begin{tabular}{@{}llcc@{}}
\toprule
Data set & Arm & Recall & CPD \\
\midrule
MNIST & base & $0.431 \pm 0.000$ & $0.636 \pm 0.000$ \\
 & $L=2$ & $0.431 \pm 0.000$ & $0.640 \pm 0.000$ \\
 & $L=8$ & $0.430 \pm 0.000$ & $0.640 \pm 0.002$ \\
 & no sketch & $0.431 \pm 0.001$ & $0.641 \pm 0.002$ \\
 & +NLL & $0.431 \pm 0.000$ & $0.635 \pm 0.004$ \\
 & no Fourier branch & $0.431 \pm 0.001$ & $0.637 \pm 0.002$ \\
 & scalar gate & $0.432 \pm 0.001$ & $0.641 \pm 0.002$ \\
 & 800 steps & $0.427 \pm 0.000$ & $0.639 \pm 0.002$ \\
\midrule
Fashion-MNIST & base & $0.434 \pm 0.001$ & $0.823 \pm 0.001$ \\
 & $L=2$ & $0.435 \pm 0.001$ & $0.823 \pm 0.001$ \\
 & $L=8$ & $0.433 \pm 0.001$ & $0.826 \pm 0.001$ \\
 & no sketch & $0.435 \pm 0.001$ & $0.823 \pm 0.000$ \\
 & +NLL & $0.434 \pm 0.001$ & $0.824 \pm 0.000$ \\
 & no Fourier branch & $0.434 \pm 0.001$ & $0.823 \pm 0.001$ \\
 & scalar gate & $0.433 \pm 0.001$ & $0.823 \pm 0.001$ \\
 & 800 steps & $0.431 \pm 0.001$ & $0.821 \pm 0.001$ \\
\midrule
paul15 & base & $0.168 \pm 0.001$ & $0.730 \pm 0.002$ \\
 & $L=2$ & $0.166 \pm 0.001$ & $0.733 \pm 0.003$ \\
 & $L=8$ & $0.167 \pm 0.001$ & $0.728 \pm 0.001$ \\
 & +NLL & $0.165 \pm 0.001$ & $0.719 \pm 0.000$ \\
 & no Fourier branch & $0.167 \pm 0.001$ & $0.730 \pm 0.002$ \\
 & scalar gate & $0.167 \pm 0.000$ & $0.730 \pm 0.001$ \\
 & 800 steps & $0.165 \pm 0.001$ & $0.733 \pm 0.001$ \\
\midrule
Schneider50k & base & $0.516 \pm 0.000$ & $0.287 \pm 0.004$ \\
 & $L=2$ & $0.515 \pm 0.001$ & $0.289 \pm 0.003$ \\
 & $L=8$ & $0.514 \pm 0.002$ & $0.294 \pm 0.006$ \\
 & no sketch & $0.517 \pm 0.001$ & $0.282 \pm 0.008$ \\
 & +NLL & $0.515 \pm 0.001$ & $0.288 \pm 0.006$ \\
 & no Fourier branch & $0.515 \pm 0.001$ & $0.286 \pm 0.007$ \\
 & scalar gate & $0.517 \pm 0.000$ & $0.282 \pm 0.004$ \\
 & 800 steps & $0.511 \pm 0.001$ & $0.268 \pm 0.007$ \\
\botrule
\end{tabular}}
\end{table}

\begin{table}[h!]
\caption{\textbf{Scores on the dimensionality reduction task of OpenProblems per metric.} Means over the five data sets of Table~\ref{tab:openproblems}, each metric scaled per data set between the worst and the best of three control methods, where a missing value counts as zero. Clustering is adjusted Rand index (ARI) and normalised mutual information (NMI) of Leiden clusters of the embedding against the cell types. Co-ranking is continuity (Cont.), trustworthiness (Trust.), $Q_{NX}$ (QNX) and the local continuity meta-criterion (LCMC) at $k=30$, of which QNX and LCMC differ by a constant and so scale alike, and the area under the $Q_{NX}$ curve (AUC) with its means up to ($Q_\mathrm{loc}$) and beyond ($Q_\mathrm{glob}$) the neighbourhood size of the largest LCMC. Density (Dens.) is Pearson correlation of the logarithms of the local radii of densMAP~\cite{Narayan2021}, input against embedding. Distance correlations are Spearman correlations of the distances from waypoint cells to the cell-type centroids (Centr.), between centroids (Label) and between waypoint cells (Waypt.). Spectral (Spec.) is Spearman correlation with the distances of a spectral embedding of the input. The methods are NeuralEE~\cite{Xiong2020}, PHATE~\cite{Moon2017PHATEAD}, ivis~\cite{Szubert2019StructurepreservingVO}, t-SNE~\cite{Maaten2008VisualizingDU}, PyMDE~\cite{Agrawal2021MinimumDistortionE}, UMAP~\cite{McInnes2018UMAPUM}, LMDS~\cite{Saelens2019}, densMAP, diffusion maps~\cite{Coifman2006} and SIMLR~\cite{Wang2017}.}
\label{tab:openproblems-metrics}
{\footnotesize\setlength{\tabcolsep}{2.4pt}\def\rh#1{\rotatebox{90}{#1}}
\begin{tabular}{@{}lcccccccccccccc@{}}
\toprule
 & \multicolumn{2}{c}{Clustering} & \multicolumn{7}{c}{Co-ranking} & & \multicolumn{3}{c}{Distance corr.} & \\
\cmidrule(lr){2-3}\cmidrule(lr){4-10}\cmidrule(lr){12-14}
 & \rh{ARI} & \rh{NMI} & \rh{Cont.} & \rh{Trust.} & \rh{QNX} & \rh{LCMC} & \rh{AUC} & \rh{$Q_\mathrm{loc}$} & \rh{$Q_\mathrm{glob}$} & \rh{Dens.} & \rh{Centr.} & \rh{Label} & \rh{Waypt.} & \rh{Spec.} \\
\midrule
FloDR & 0.87 & 1.00 & 0.80 & 0.78 & 0.36 & 0.36 & 0.50 & 0.39 & 0.41 & 0.43 & 0.60 & 0.58 & 0.65 & 0.31 \\
FloDR, density term & 0.83 & 1.00 & 0.81 & 0.78 & 0.36 & 0.36 & 0.52 & 0.37 & 0.41 & 0.87 & 0.61 & 0.58 & 0.65 & 0.31 \\
\midrule
NeuralEE & 0.89 & 0.98 & 0.71 & 0.67 & 0.30 & 0.30 & 0.38 & 0.33 & 0.32 & 0.43 & 0.26 & 0.32 & 0.32 & 0.28 \\
PHATE & 0.80 & 0.98 & 0.79 & 0.82 & 0.34 & 0.34 & 0.43 & 0.38 & 0.32 & 0.09 & 0.30 & 0.27 & 0.37 & 0.25 \\
ivis & 0.93 & 0.99 & 0.71 & 0.67 & 0.31 & 0.31 & 0.39 & 0.33 & 0.30 & 0.38 & 0.29 & 0.27 & 0.32 & 0.12 \\
t-SNE & 0.76 & 0.99 & 0.68 & 0.79 & 0.35 & 0.35 & 0.40 & 0.35 & 0.29 & 0.14 & 0.25 & 0.34 & 0.29 & 0.29 \\
PyMDE & 0.74 & 0.97 & 0.58 & 0.62 & 0.32 & 0.32 & 0.36 & 0.34 & 0.31 & 0.13 & 0.33 & 0.52 & 0.31 & 0.23 \\
PCA & 0.55 & 0.69 & 0.69 & 0.63 & 0.24 & 0.24 & 0.38 & 0.35 & 0.33 & 0.27 & 0.37 & 0.60 & 0.34 & 0.29 \\
UMAP & 0.88 & 1.00 & 0.57 & 0.58 & 0.31 & 0.31 & 0.30 & 0.27 & 0.15 & 0.03 & 0.20 & 0.34 & 0.23 & 0.25 \\
LMDS & 0.49 & 0.59 & 0.52 & 0.51 & 0.22 & 0.22 & 0.32 & 0.27 & 0.23 & 0.28 & 0.28 & 0.42 & 0.26 & 0.27 \\
densMAP & 0.57 & 0.66 & 0.28 & 0.24 & 0.12 & 0.12 & 0.09 & 0.09 & 0.05 & 0.49 & 0.28 & 0.40 & 0.27 & 0.26 \\
Diffusion map & 0.32 & 0.42 & 0.21 & 0.30 & 0.09 & 0.09 & 0.06 & 0.08 & 0.00 & 0.08 & 0.02 & 0.16 & 0.02 & 0.27 \\
SIMLR & 0.00 & 0.00 & 0.19 & 0.19 & 0.09 & 0.09 & 0.08 & 0.09 & 0.05 & 0.15 & 0.11 & 0.02 & 0.10 & 0.20 \\
\botrule
\end{tabular}}
\end{table}

\clearpage
\section*{Supplementary Information}

\setcounter{table}{0}
\renewcommand{\tablename}{Supplementary Table}
\renewcommand{\theHtable}{S\arabic{table}}

\begin{table}[h!]
\caption{\textbf{FloDR against DREAMS.} Neighbourhood recall at $k=15$ and CPD on the held-out pairs, means over three seeds, for FloDR at its default ($w=2$) and for DREAMS over its regularisation strength $\lambda$, where $\lambda=0$ is t-SNE and $\lambda=0.15$ is the default. Inputs, metrics, and seeds are those used in Fig.~3 in the main text.}
{\setlength{\tabcolsep}{4pt}
\begin{tabular}{@{}lccccccc@{}}
\toprule
& & \multicolumn{6}{c}{DREAMS, $\lambda$} \\
\cmidrule(l){3-8}
 & FloDR & 0 & 0.05 & 0.15 & 0.3 & 0.5 & 0.8 \\
\midrule
\multicolumn{8}{@{}l}{\textit{Recall}} \\
MNIST & 0.431 & 0.436 & 0.435 & 0.431 & 0.410 & 0.300 & 0.105 \\
Fashion-MNIST & 0.434 & 0.441 & 0.442 & 0.439 & 0.424 & 0.337 & 0.164 \\
paul15 & 0.168 & 0.170 & 0.171 & 0.155 & 0.124 & 0.092 & 0.025 \\
Schneider50k & 0.516 & 0.497 & 0.502 & 0.495 & 0.451 & 0.331 & 0.135 \\
Bone marrow & 0.187 & 0.280 & 0.229 & 0.159 & 0.117 & 0.054 & 0.010 \\
Cerebellum & 0.156 & 0.234 & 0.196 & 0.145 & 0.095 & 0.037 & 0.006 \\
\textit{Arabidopsis} seed & 0.089 & 0.150 & 0.116 & 0.059 & 0.037 & 0.019 & 0.005 \\
\midrule
\multicolumn{8}{@{}l}{\textit{CPD}} \\
MNIST & 0.636 & 0.410 & 0.408 & 0.407 & 0.423 & 0.458 & 0.484 \\
Fashion-MNIST & 0.823 & 0.691 & 0.735 & 0.808 & 0.833 & 0.861 & 0.878 \\
paul15 & 0.730 & 0.295 & 0.317 & 0.367 & 0.415 & 0.505 & 0.576 \\
Schneider50k & 0.287 & 0.203 & 0.199 & 0.174 & 0.135 & 0.122 & 0.108 \\
Bone marrow & 0.836 & 0.240 & 0.358 & 0.468 & 0.515 & 0.550 & 0.565 \\
Cerebellum & 0.566 & 0.191 & 0.242 & 0.303 & 0.324 & 0.372 & 0.403 \\
\textit{Arabidopsis} seed & 0.709 & 0.407 & 0.415 & 0.450 & 0.473 & 0.518 & 0.530 \\
\botrule
\end{tabular}}
\end{table}

\begin{table}[h!]
\caption{\textbf{Further settings of three baselines.} Neighbourhood recall at $k=15$ and CPD on the held-out pairs, means over three seeds, for FloDR, UMAP, openTSNE, and PaCMAP at their defaults, and for 16 further settings that vary UMAP's minimum distance and number of negative samples per edge, openTSNE's exaggeration and PaCMAP's ratios of mid-near and further pairs, with the default of each in brackets. Inputs, metrics, and seeds are those used in Fig.~3 in the main text, and Seed is the \textit{Arabidopsis} seed atlas. Values above that of FloDR are in bold.}
{\footnotesize\setlength{\tabcolsep}{3pt}
\begin{tabular}{@{}lccccccc@{}}
\toprule
 & MNIST & Fashion-MNIST & paul15 & Schneider50k & Bone marrow & Cerebellum & Seed \\
\midrule
\multicolumn{8}{@{}l}{\textit{Recall}} \\
FloDR, $w=2$ & $0.431$ & $0.434$ & $0.168$ & $0.516$ & $0.187$ & $0.156$ & $0.089$ \\
\addlinespace[2pt]
UMAP, default ($0.1$, 5) & $0.339$ & $0.350$ & $0.102$ & $0.437$ & $0.119$ & $0.092$ & $0.035$ \\
openTSNE, default (1) & $\mathbf{0.435}$ & $\mathbf{0.442}$ & $\mathbf{0.169}$ & $0.499$ & $\mathbf{0.279}$ & $\mathbf{0.234}$ & $\mathbf{0.150}$ \\
PaCMAP, default ($0.5$, 2) & $0.274$ & $0.291$ & $0.072$ & $0.353$ & $0.096$ & $0.074$ & $0.026$ \\
\addlinespace[2pt]
UMAP, 0.001, 5 & $0.349$ & $0.356$ & $0.114$ & $0.433$ & $0.121$ & $0.097$ & $0.040$ \\
UMAP, 0.001, 15 & $0.391$ & $0.395$ & $0.147$ & $0.468$ & $0.143$ & $0.118$ & $0.055$ \\
UMAP, 0.01, 5 & $0.348$ & $0.358$ & $0.113$ & $0.436$ & $0.121$ & $0.097$ & $0.039$ \\
UMAP, 0.01, 15 & $0.390$ & $0.393$ & $0.147$ & $0.468$ & $0.142$ & $0.117$ & $0.055$ \\
UMAP, 0.1, 15 & $0.383$ & $0.389$ & $0.136$ & $0.471$ & $0.137$ & $0.112$ & $0.048$ \\
UMAP, 0.5, 5 & $0.281$ & $0.296$ & $0.075$ & $0.371$ & $0.097$ & $0.064$ & $0.022$ \\
UMAP, 0.5, 15 & $0.324$ & $0.340$ & $0.096$ & $0.431$ & $0.109$ & $0.080$ & $0.030$ \\
openTSNE, exaggeration 2 & $0.353$ & $0.371$ & $0.066$ & $0.459$ & $0.160$ & $0.130$ & $0.032$ \\
openTSNE, exaggeration 4 & $0.259$ & $0.293$ & $0.048$ & $0.337$ & $0.117$ & $0.091$ & $0.021$ \\
openTSNE, exaggeration 8 & $0.191$ & $0.259$ & $0.010$ & $0.228$ & $0.099$ & $0.069$ & $0.018$ \\
openTSNE, exaggeration 12 & $0.167$ & $0.246$ & $0.010$ & $0.203$ & $0.093$ & $0.062$ & $0.017$ \\
PaCMAP, 0.5, 1 & $0.239$ & $0.273$ & $0.063$ & $0.330$ & $0.089$ & $0.064$ & $0.022$ \\
PaCMAP, 0.5, 4 & $0.302$ & $0.318$ & $0.084$ & $0.377$ & $0.103$ & $0.085$ & $0.030$ \\
PaCMAP, 2, 1 & $0.240$ & $0.273$ & $0.064$ & $0.330$ & $0.089$ & $0.065$ & $0.022$ \\
PaCMAP, 2, 2 & $0.273$ & $0.291$ & $0.072$ & $0.354$ & $0.094$ & $0.074$ & $0.025$ \\
PaCMAP, 2, 4 & $0.302$ & $0.317$ & $0.085$ & $0.375$ & $0.103$ & $0.084$ & $0.030$ \\
\midrule
\multicolumn{8}{@{}l}{\textit{CPD}} \\
FloDR, $w=2$ & $0.636$ & $0.823$ & $0.730$ & $0.287$ & $0.836$ & $0.566$ & $0.709$ \\
\addlinespace[2pt]
UMAP, default ($0.1$, 5) & $0.363$ & $0.604$ & $0.049$ & $0.189$ & $0.220$ & $0.123$ & $0.495$ \\
openTSNE, default (1) & $0.415$ & $0.699$ & $0.327$ & $0.199$ & $0.301$ & $0.187$ & $0.404$ \\
PaCMAP, default ($0.5$, 2) & $0.250$ & $0.611$ & $0.050$ & $0.105$ & $0.377$ & $0.202$ & $0.567$ \\
\addlinespace[2pt]
UMAP, 0.001, 5 & $0.348$ & $0.586$ & $0.055$ & $0.200$ & $0.195$ & $0.123$ & $0.494$ \\
UMAP, 0.001, 15 & $0.371$ & $0.589$ & $0.072$ & $0.207$ & $0.148$ & $0.166$ & $0.460$ \\
UMAP, 0.01, 5 & $0.357$ & $0.585$ & $0.062$ & $0.202$ & $0.207$ & $0.120$ & $0.501$ \\
UMAP, 0.01, 15 & $0.339$ & $0.595$ & $0.078$ & $0.203$ & $0.144$ & $0.161$ & $0.460$ \\
UMAP, 0.1, 15 & $0.382$ & $0.590$ & $0.067$ & $0.202$ & $0.153$ & $0.163$ & $0.457$ \\
UMAP, 0.5, 5 & $0.383$ & $0.686$ & $0.044$ & $0.189$ & $0.238$ & $0.154$ & $0.540$ \\
UMAP, 0.5, 15 & $0.371$ & $0.662$ & $0.059$ & $0.195$ & $0.191$ & $0.126$ & $0.490$ \\
openTSNE, exaggeration 2 & $0.410$ & $0.628$ & $0.206$ & $0.207$ & $0.291$ & $0.156$ & $0.543$ \\
openTSNE, exaggeration 4 & $0.412$ & $0.632$ & $0.064$ & $0.213$ & $0.299$ & $0.131$ & $0.661$ \\
openTSNE, exaggeration 8 & $0.452$ & $0.677$ & $-0.305$ & $0.222$ & $0.292$ & $0.100$ & $\mathbf{0.721}$ \\
openTSNE, exaggeration 12 & $0.450$ & $0.678$ & $-0.269$ & $0.212$ & $0.303$ & $0.083$ & $\mathbf{0.727}$ \\
PaCMAP, 0.5, 1 & $0.194$ & $0.609$ & $0.016$ & $0.113$ & $0.310$ & $0.166$ & $0.655$ \\
PaCMAP, 0.5, 4 & $0.294$ & $0.607$ & $0.117$ & $0.099$ & $0.407$ & $0.242$ & $0.471$ \\
PaCMAP, 2, 1 & $0.176$ & $0.617$ & $0.017$ & $0.107$ & $0.422$ & $0.221$ & $0.657$ \\
PaCMAP, 2, 2 & $0.282$ & $0.626$ & $0.050$ & $0.105$ & $0.416$ & $0.252$ & $0.595$ \\
PaCMAP, 2, 4 & $0.288$ & $0.630$ & $0.102$ & $0.110$ & $0.408$ & $0.270$ & $0.472$ \\
\botrule
\end{tabular}}
\end{table}

\clearpage

\begin{table}[h!]
\caption{\textbf{Scores on version 1 of the dimensionality reduction task of OpenProblems.} Its leaderboard is the one shown on the OpenProblems website. Each metric is scaled per data set between the worst and the best of three baselines, clipped to $[0,1]$ and averaged over data sets and metrics, where a missing value counts as zero, same as on the website. On zebrafish, the published results have four of the ten metrics for every method, and we scored FloDR on the same four. FloDR was fitted on $\min(500, n-1)$ principal components of the $\log$ CP10k matrix and scored with the metric functions of the task (Methods), while the other rows are the 23 published variants under their published names. The data sets are mouse myeloid progenitors (Myeloid, 660 cells)~\cite{Olsson2016}, mouse HSPC ($1{,}920$)~\cite{Nestorowa2016}, peripheral blood mononuclear cells from 10x Genomics (PBMC, $5{,}247$), and zebrafish embryos ($26{,}022$)~\cite{Wagner2018}.}
{\footnotesize
\begin{tabular}{@{}lccccc@{}}
\toprule
Method & Myeloid & HSPC & PBMC & Zebrafish & Mean \\
\midrule
FloDR & 0.525 & 0.355 & 0.506 & 0.214 & 0.400 \\
FloDR, density term & 0.572 & 0.377 & 0.524 & 0.254 & 0.432 \\
\midrule
densMAP (logCP10k) & 0.506 & 0.231 & 0.465 & 0.219 & 0.355 \\
densMAP PCA (logCP10k) & 0.413 & 0.227 & 0.439 & 0.179 & 0.314 \\
t-SNE (logCP10k) & 0.421 & 0.192 & 0.393 & 0.153 & 0.290 \\
PyMDE Preserve Distances (logCP10k) & 0.292 & 0.208 & 0.442 & 0.183 & 0.281 \\
densMAP PCA (logCP10k, 1kHVG) & 0.444 & 0.120 & 0.377 & 0.182 & 0.281 \\
NeuralEE (CPU) (logCP10k, 1kHVG) & 0.473 & 0.135 & 0.369 & 0.142 & 0.280 \\
PyMDE Preserve Neighbors (logCP10k) & 0.440 & 0.154 & 0.365 & 0.145 & 0.276 \\
UMAP (logCP10k) & 0.432 & 0.162 & 0.374 & 0.135 & 0.276 \\
densMAP (logCP10k, 1kHVG) & 0.383 & 0.132 & 0.410 & 0.173 & 0.275 \\
PyMDE Preserve Neighbors (logCP10k, 1kHVG) & 0.447 & 0.136 & 0.355 & 0.154 & 0.273 \\
t-SNE (logCP10k, 1kHVG) & 0.461 & 0.143 & 0.338 & 0.143 & 0.271 \\
UMAP PCA (logCP10k) & 0.419 & 0.141 & 0.381 & 0.143 & 0.271 \\
PHATE (logCP10k, 1kHVG) & 0.429 & 0.099 & 0.351 & 0.162 & 0.260 \\
UMAP PCA (logCP10k, 1kHVG) & 0.433 & 0.120 & 0.329 & 0.153 & 0.259 \\
PHATE ($\gamma=0$) & 0.407 & 0.092 & 0.371 & 0.155 & 0.256 \\
PCA (logCP10k, 1kHVG) & 0.482 & 0.082 & 0.332 & 0.120 & 0.254 \\
PHATE (logCP10k) & 0.409 & 0.106 & 0.371 & 0.127 & 0.253 \\
NeuralEE (CPU) (Default) & 0.438 & 0.107 & 0.342 & 0.123 & 0.252 \\
PCA (logCP10k) & 0.330 & 0.101 & 0.424 & 0.143 & 0.249 \\
PHATE (default) & 0.414 & 0.091 & 0.347 & 0.143 & 0.248 \\
UMAP (logCP10k, 1kHVG) & 0.358 & 0.117 & 0.348 & 0.144 & 0.242 \\
PyMDE Preserve Distances (logCP10k, 1kHVG) & 0.351 & 0.122 & 0.349 & 0.136 & 0.240 \\
Diffusion maps & 0.322 & 0.042 & 0.336 & 0.136 & 0.209 \\
\botrule
\end{tabular}}
\end{table}

\clearpage

{\footnotesize\setlength{\tabcolsep}{5pt}
\begin{longtable}{@{}p{0.31\linewidth}p{0.15\linewidth}p{0.15\linewidth}p{0.15\linewidth}p{0.15\linewidth}@{}}
\caption{\textbf{Every readout of the tests against unseen measurements.} Means over the seeds of each map. A letter gives the verdict of FloDR against the map of that row by the decision rule. W is a win, T is avtie, and L is a loss (Methods). $^{*}$ is the primary readout. $^{\circ}$ supports no claim, as the input did not exceed UMAP by $0.03$ (in g and h, the transform of UMAP). Readouts without letters are reported only. } \\
\toprule
\endfirsthead
\toprule
\endhead
\addlinespace[4pt]\multicolumn{5}{@{}p{\linewidth}@{}}{\textbf{a}, Genotypes of the HGDP-CEPH panel, five seeds. Agreement of map distances between populations with distances between sampling sites, on the map and for the seven held-out populations, and the share of held-out individuals whose nearest population on the map is of their region.} \\
 & Map$^{*}$ & Held out$^{*}$ & Region &  \\
\midrule
FloDR & $0.768$ & $0.775$ & $0.857$ &  \\
FloDR, forward pass & -- & $0.776$\textsuperscript{T} & $0.988$ &  \\
FloDR, heat start & $0.719$\textsuperscript{W} & $0.603$\textsuperscript{T} & $0.857$ &  \\
FloDR, $w=0$ & $0.554$\textsuperscript{W} & $0.654$\textsuperscript{W} & $0.857$ &  \\
UMAP & $0.465$\textsuperscript{W} & $0.475$\textsuperscript{W} & $0.857$ &  \\
openTSNE & $0.695$\textsuperscript{W} & $0.736$\textsuperscript{T} & $0.857$ &  \\
PHATE & $0.214$\textsuperscript{W} & $0.183$\textsuperscript{W} & $0.857$ &  \\
Parametric UMAP & $0.685$\textsuperscript{W} & $0.652$\textsuperscript{T} & $0.896$ &  \\
PaCMAP & $0.638$\textsuperscript{W} & -- & -- &  \\
DREAMS & $0.630$\textsuperscript{W} & -- & -- &  \\
PCA, two components & $0.621$\textsuperscript{W} & $0.643$\textsuperscript{W} & $0.914$ &  \\
Non-metric MDS, one fit & $0.835$\textsuperscript{L} & -- & -- &  \\
kNN transfer, map of FloDR & -- & $0.785$\textsuperscript{T} & $0.857$ &  \\
kNN transfer, map of UMAP & -- & $0.476$\textsuperscript{W} & $0.821$ &  \\
Input, 50 components & $0.850$ & $0.842$ & $1.000$ &  \\
\midrule
\addlinespace[4pt]\multicolumn{5}{@{}p{\linewidth}@{}}{\textbf{b}, Cortical depth, SEA-AD, three seeds. Agreement of map distances between depth bins with their separation, for the 19 donors of the map and the 8 placed donors, and the order by depth within subclasses.} \\
 & Map$^{*}$$^{\circ}$ & Placed$^{*}$$^{\circ}$ & Within subclass, map & Within subclass, placed \\
\midrule
FloDR & $0.912$ & $0.904$ & $0.112$ & $0.138$ \\
FloDR, forward pass & -- & $0.902$\textsuperscript{T} & $0.112$ & $0.138$ \\
FloDR, heat start & $0.866$\textsuperscript{W} & $0.845$\textsuperscript{W} & $0.105$ & $0.128$ \\
FloDR, $w=0$ & $0.864$\textsuperscript{W} & $0.872$\textsuperscript{W} & $0.129$ & $0.155$ \\
UMAP & $0.905$\textsuperscript{T} & $0.919$\textsuperscript{T} & $0.116$ & $0.146$ \\
openTSNE & $0.914$\textsuperscript{T} & $0.900$\textsuperscript{T} & $0.129$ & $0.152$ \\
PHATE & $0.878$\textsuperscript{W} & $0.880$\textsuperscript{T} & $0.114$ & $0.115$ \\
Parametric UMAP & $0.843$\textsuperscript{W} & $0.854$\textsuperscript{W} & $0.148$ & $0.176$ \\
PaCMAP & $0.879$\textsuperscript{W} & -- & $0.120$ & -- \\
DREAMS & $0.823$\textsuperscript{W} & -- & $0.092$ & -- \\
PCA, two components & $0.699$\textsuperscript{W} & $0.670$\textsuperscript{W} & $0.097$ & $0.113$ \\
kNN transfer, map of FloDR & -- & $0.902$\textsuperscript{T} & $0.112$ & $0.137$ \\
kNN transfer, map of UMAP & -- & $0.915$\textsuperscript{T} & $0.116$ & $0.143$ \\
Input, 50 components & $0.863$ & $0.867$ & $0.106$ & $0.129$ \\
\midrule
\addlinespace[4pt]\multicolumn{5}{@{}p{\linewidth}@{}}{\textbf{c}, LARRY, five seeds. Coupling of cell types within clones, fate of the clone of a progenitor (neutrophil against monocyte, and the mean over all pairs of fates), and sister cells against matched non-sisters. The layout published with the data~\cite{Weinreb2020} is a reference without verdicts, as is the input.} \\
 & Coupling$^{*}$$^{\circ}$ & Fate$^{*}$ & Fate, all pairs & Sisters \\
\midrule
FloDR & $0.512$ & $0.755$ & $0.821$ & $0.638$ \\
FloDR, $w=0$ & $0.525$\textsuperscript{T} & $0.752$\textsuperscript{T} & $0.865$ & $0.655$\textsuperscript{T} \\
UMAP & $0.386$\textsuperscript{W} & $0.471$\textsuperscript{W} & $0.807$ & $0.661$\textsuperscript{L} \\
openTSNE & $0.432$\textsuperscript{T} & $0.480$\textsuperscript{W} & $0.796$ & $0.662$\textsuperscript{L} \\
PaCMAP & $0.297$\textsuperscript{W} & $0.491$\textsuperscript{W} & $0.805$ & $0.662$\textsuperscript{L} \\
densMAP & $0.309$\textsuperscript{W} & $0.466$\textsuperscript{W} & $0.769$ & $0.650$\textsuperscript{T} \\
DREAMS & $0.312$\textsuperscript{W} & $0.392$\textsuperscript{W} & $0.802$ & $0.628$\textsuperscript{T} \\
PHATE & $0.722$\textsuperscript{L} & $0.816$\textsuperscript{L} & $0.762$ & $0.664$\textsuperscript{L} \\
PCA, two components & $0.142$\textsuperscript{W} & $0.385$\textsuperscript{W} & $0.766$ & $0.597$\textsuperscript{W} \\
Input, 50 components & $0.289$ & $0.735$ & $0.893$ & $0.616$ \\
Layout published with the data & $0.693$ & $0.756$ & $0.888$ & $0.658$ \\
\midrule
\addlinespace[4pt]\multicolumn{5}{@{}p{\linewidth}@{}}{\textbf{d}, CellTag-multi, blood progenitors (LSK), three seeds.  A further arm, FloDR on the heat-augmented input, maps the input together with three times its 50 heat-geodesic coordinates of HeatGeo~\cite{Huguet2023heat} and takes its neighbours and the distances its ordinal term ranks from those coordinates. The verdicts use $1\,000$ replicates, that against the further arm $2\,000$.} \\
 & Fate$^{*}$ & Fate, three siblings & Sisters & Coupling \\
\midrule
FloDR & $0.773$ & $0.805$ & $0.630$ & $0.663$ \\
FloDR, heat-augmented input & $0.567$\textsuperscript{W} & $0.670$\textsuperscript{W} & $0.646$\textsuperscript{T} & $0.565$\textsuperscript{W} \\
UMAP & $0.405$\textsuperscript{W} & $0.342$\textsuperscript{W} & $0.629$\textsuperscript{T} & $0.691$\textsuperscript{T} \\
openTSNE & $0.632$\textsuperscript{W} & $0.704$\textsuperscript{W} & $0.643$\textsuperscript{T} & $0.721$\textsuperscript{T} \\
PaCMAP & $0.486$\textsuperscript{W} & $0.405$\textsuperscript{W} & $0.632$\textsuperscript{T} & $0.713$\textsuperscript{L} \\
densMAP & $0.399$\textsuperscript{W} & $0.328$\textsuperscript{W} & $0.622$\textsuperscript{T} & $0.716$\textsuperscript{L} \\
DREAMS & $0.673$\textsuperscript{W} & $0.722$\textsuperscript{W} & $0.635$\textsuperscript{T} & $0.463$\textsuperscript{W} \\
PHATE & $0.603$\textsuperscript{W} & $0.613$\textsuperscript{W} & $0.666$\textsuperscript{L} & $0.526$\textsuperscript{T} \\
PCA, two components & $0.669$\textsuperscript{W} & $0.725$\textsuperscript{W} & $0.633$\textsuperscript{T} & $0.398$\textsuperscript{W} \\
Input, 50 components & $0.581$ & $0.669$ & $0.644$ & $0.351$ \\
\midrule
\addlinespace[4pt]\multicolumn{5}{@{}p{\linewidth}@{}}{\textbf{e}, CellTag-multi, reprogramming, two replicates of three seeds each, planned and scored as d.} \\
 & Fate$^{*}$$^{\circ}$ & Fate, three siblings & Sisters & Coupling \\
\midrule
FloDR & $0.833$ & $0.832$ & $0.600$ & $0.457$ \\
FloDR, heat-augmented input & $0.841$\textsuperscript{T} & $0.845$\textsuperscript{T} & $0.600$\textsuperscript{T} & $0.462$\textsuperscript{T} \\
UMAP & $0.838$\textsuperscript{T} & $0.851$\textsuperscript{T} & $0.577$\textsuperscript{W} & $0.177$\textsuperscript{T} \\
openTSNE & $0.809$\textsuperscript{W} & $0.784$\textsuperscript{W} & $0.583$\textsuperscript{T} & $0.279$\textsuperscript{T} \\
PaCMAP & $0.796$\textsuperscript{W} & $0.806$\textsuperscript{T} & $0.576$\textsuperscript{W} & $-0.055$\textsuperscript{W} \\
densMAP & $0.819$\textsuperscript{T} & $0.794$\textsuperscript{T} & $0.587$\textsuperscript{T} & $0.070$\textsuperscript{W} \\
DREAMS & $0.840$\textsuperscript{T} & $0.826$\textsuperscript{T} & $0.585$\textsuperscript{T} & $0.711$\textsuperscript{L} \\
PHATE & $0.803$\textsuperscript{T} & $0.815$\textsuperscript{T} & $0.583$\textsuperscript{T} & $0.728$\textsuperscript{L} \\
PCA, two components & $0.847$\textsuperscript{T} & $0.840$\textsuperscript{T} & $0.594$\textsuperscript{T} & $0.843$\textsuperscript{L} \\
Input, 50 components & $0.861$ & $0.876$ & $0.686$ & $0.473$ \\
\midrule
\addlinespace[4pt]\multicolumn{5}{@{}p{\linewidth}@{}}{\textbf{f}, CITE-seq, 12 batches, three seeds. Agreement of map distances with distances in protein levels, between cell types and between cells, with a Wilcoxon test over the batches.} \\
 & Cell types & Cells &  &  \\
\midrule
FloDR & $0.495$ & $0.569$ &  &  \\
FloDR, $w=0$ & $0.545$\textsuperscript{T} & $0.577$\textsuperscript{T} &  &  \\
UMAP & $0.496$\textsuperscript{T} & $0.581$\textsuperscript{T} &  &  \\
openTSNE & $0.484$\textsuperscript{T} & $0.542$\textsuperscript{T} &  &  \\
PaCMAP & $0.534$\textsuperscript{T} & $0.601$\textsuperscript{T} &  &  \\
densMAP & $0.496$\textsuperscript{T} & $0.587$\textsuperscript{T} &  &  \\
DREAMS & $0.455$\textsuperscript{T} & $0.586$\textsuperscript{T} &  &  \\
PHATE & $0.479$\textsuperscript{T} & $0.565$\textsuperscript{T} &  &  \\
PCA, two components & $0.446$\textsuperscript{T} & $0.580$\textsuperscript{T} &  &  \\
Input, 50 components & $0.487$ & $0.550$ &  &  \\
\midrule
\addlinespace[4pt]\multicolumn{5}{@{}p{\linewidth}@{}}{\textbf{g}, Placement of progenitors on a map fitted without them, FloDR with the heat start, LARRY, five seeds. Fate reads the placed cells, coupling and sisters the map.} \\
 & Fate$^{*}$$^{\circ}$ & Fate, all pairs & Coupling & Sisters \\
\midrule
FloDR, heat start, optimised step & $0.769$ & $0.880$ & $0.675$ & $0.634$ \\
FloDR, forward pass & $0.775$\textsuperscript{T} & $0.875$\textsuperscript{T} & $0.675$\textsuperscript{T} & $0.634$\textsuperscript{T} \\
UMAP & $0.748$\textsuperscript{W} & $0.846$\textsuperscript{W} & $0.423$\textsuperscript{W} & $0.656$\textsuperscript{L} \\
openTSNE & $0.761$\textsuperscript{T} & $0.892$\textsuperscript{T} & $0.637$\textsuperscript{T} & $0.649$\textsuperscript{T} \\
PHATE & $0.785$\textsuperscript{T} & $0.790$\textsuperscript{W} & $0.725$\textsuperscript{L} & $0.658$\textsuperscript{L} \\
Parametric UMAP & $0.689$\textsuperscript{T} & $0.816$\textsuperscript{W} & $0.591$\textsuperscript{T} & $0.650$\textsuperscript{T} \\
kNN transfer, map of FloDR & $0.776$\textsuperscript{T} & $0.875$\textsuperscript{T} & $0.675$\textsuperscript{T} & $0.634$\textsuperscript{T} \\
kNN transfer, map of UMAP & $0.747$\textsuperscript{W} & $0.850$\textsuperscript{W} & $0.423$\textsuperscript{W} & $0.656$\textsuperscript{L} \\
Input, 50 components & $0.745$ & $0.888$ & -- & -- \\
\midrule
\addlinespace[4pt]\multicolumn{5}{@{}p{\linewidth}@{}}{\textbf{h}, The same on the LSK arm of CellTag-multi, three seeds.} \\
 & Fate$^{*}$$^{\circ}$ & Fate, three siblings & Coupling & Sisters \\
\midrule
FloDR, heat start, optimised step & $0.682$ & $0.606$ & $0.550$ & $0.631$ \\
FloDR, forward pass & $0.738$\textsuperscript{L} & $0.700$\textsuperscript{L} & $0.550$\textsuperscript{T} & $0.631$\textsuperscript{T} \\
UMAP & $0.736$\textsuperscript{T} & $0.679$\textsuperscript{T} & $0.694$\textsuperscript{L} & $0.629$\textsuperscript{T} \\
openTSNE & $0.677$\textsuperscript{T} & $0.723$\textsuperscript{L} & $0.754$\textsuperscript{L} & $0.640$\textsuperscript{T} \\
PHATE & $0.795$\textsuperscript{L} & $0.753$\textsuperscript{L} & $0.592$\textsuperscript{T} & $0.667$\textsuperscript{L} \\
Parametric UMAP & $0.388$\textsuperscript{T} & $0.418$\textsuperscript{T} & $0.598$\textsuperscript{T} & $0.629$\textsuperscript{T} \\
kNN transfer, map of FloDR & $0.786$\textsuperscript{L} & $0.741$\textsuperscript{L} & $0.550$\textsuperscript{T} & $0.631$\textsuperscript{T} \\
kNN transfer, map of UMAP & $0.788$\textsuperscript{L} & $0.753$\textsuperscript{L} & $0.694$\textsuperscript{L} & $0.629$\textsuperscript{T} \\
Input, 50 components & $0.626$ & $0.717$ & -- & -- \\
\bottomrule
\end{longtable}}

\begin{table}[h!]
\caption{\textbf{Neighbours and distance ranks on the inputs of the tests against unseen measurements.} Neighbourhood recall at $k=15$ (R) and CPD of each map on five of the inputs (50 principal components, not standardised), as means over the seeds of the test and, for reprogramming, over its two replicates. A dash marks a map that was not part of a test.}
{\footnotesize\setlength{\tabcolsep}{4pt}
\begin{tabular}{@{}lcccccccccc@{}}
\toprule
 & \multicolumn{2}{c}{LARRY} & \multicolumn{2}{c}{Blood progenitors} & \multicolumn{2}{c}{Reprogramming} & \multicolumn{2}{c}{Genotypes} & \multicolumn{2}{c}{Cortex} \\
\cmidrule(lr){2-3}\cmidrule(lr){4-5}\cmidrule(lr){6-7}\cmidrule(lr){8-9}\cmidrule(l){10-11}
 & R & CPD & R & CPD & R & CPD & R & CPD & R & CPD \\
\midrule
FloDR & $0.093$ & $0.918$ & $0.207$ & $0.894$ & $0.148$ & $0.780$ & $0.691$ & $0.944$ & $0.101$ & $0.617$ \\
FloDR, $w=0$ & $0.095$ & $0.699$ & -- & -- & -- & -- & $0.696$ & $0.486$ & $0.105$ & $0.448$ \\
UMAP & $0.047$ & $0.697$ & $0.110$ & $0.611$ & $0.077$ & $0.521$ & $0.633$ & $0.493$ & $0.028$ & $0.499$ \\
openTSNE & $0.190$ & $0.595$ & $0.307$ & $0.626$ & $0.247$ & $0.501$ & $0.717$ & $0.641$ & $0.182$ & $0.512$ \\
PaCMAP & $0.036$ & $0.707$ & $0.086$ & $0.589$ & $0.058$ & $0.322$ & $0.603$ & $0.618$ & $0.024$ & $0.494$ \\
densMAP & $0.020$ & $0.838$ & $0.055$ & $0.662$ & $0.034$ & $0.557$ & -- & -- & -- & -- \\
DREAMS & $0.125$ & $0.767$ & $0.283$ & $0.771$ & $0.202$ & $0.682$ & $0.718$ & $0.810$ & $0.147$ & $0.597$ \\
PHATE & $0.033$ & $0.727$ & $0.077$ & $0.661$ & $0.046$ & $0.632$ & $0.490$ & $0.052$ & $0.015$ & $0.562$ \\
Parametric UMAP & -- & -- & -- & -- & -- & -- & $0.575$ & $0.554$ & $0.015$ & $0.485$ \\
PCA, two components & $0.005$ & $0.778$ & $0.019$ & $0.791$ & $0.006$ & $0.736$ & $0.411$ & $0.819$ & $0.003$ & $0.638$ \\
\botrule
\end{tabular}}
\end{table}

\clearpage
\section*{Supplementary Note 1: Decoding map positions} 

Errors are in the units of Extended Data Fig.~3b, namely, the squared distance between decoded and held-out points, summed over the held-out points, divided by the same sum for the mean of the training points. The decodes come from the script used to create data for Extended Data Fig.~3, and the \texttt{decode} of the package returned the same inputs, within $4\times10^{-5}$ of the range of the input, on one split each of paul15 and Fashion-MNIST.

\paragraph{The zero residual.} With the residual set to zero, as in \texttt{inverse\_transform} of the released  package, the error was $5.5$ to $57.9$ on the benchmark data sets and $107$ to $1{,}301$ on the atlases, as the inverse amplifies residuals that no training point has.

\paragraph{The steps onto the fibre.} The mean of the 15 nearest training points was moved by a median of $0.047$ to $0.231$ of the median distance between a training point and its nearest neighbour in the input. This raised the error on every split by $0.001$ to $0.037$. A forward pass placed the decoded points within $10^{-2}$ of the spacing on the map from $y$. Three shortcuts through the residuals of the same neighbours did no better on the atlases. Each residual run back through the flow at $y$ and the results averaged (errors of $2.09$, $1.19$, and $0.77$ on the three atlases, compared to $0.28$, $0.48$, and $0.62$ for the decode of the Methods), the mean residual ($9.39$, $7.88$ and $9.15$), and the zero residual above. On Fashion-MNIST, averaging the residuals run back was more accurate than the decode ($0.30$ versus $0.33$, on all five splits), and on MNIST and paul15 less accurate ($0.58$ and $1.11$ versus $0.55$ and $0.93$).

\paragraph{Baselines.} We fitted Parametric UMAP with the default networks of umap-learn, three hidden layers of 100 units each for the encoder and decoder, and with the reconstruction term on the squared error, as the inputs are not scaled to $[0,1]$. The error varied most between the splits of the cerebellum atlas ($0.60$, $0.62$ and $0.81$). The inverse transform of UMAP starts from a weighted mean of as many training points around the position as the input has dimensions, which is 784 for the images and 50 for the other data sets, it then optimises this point. The result depended on the number of threads. On the split of Extended Data Fig.~3a, the error was $0.609$, $0.605$, $0.595$, and $0.596$ with 2, 4, 8, and 20 threads. For the eighth image, it returned trousers with 2 and with 8 threads, and a sneaker with 4 and with 20.

\paragraph{What a layout allows.} We decoded the same positions in two more ways. Once on the maps of FloDR and once on that of UMAP by the mean of the 15 training points nearest to the position, and also by a decoder trained on the layout as in NNInv~\cite{Espadoto2019DeepLI}. The decoder is a network with four hidden layers of $1{,}024$ rectified linear units that maps a position to an input, and is trained on the squared error with Adam at a learning rate of $10^{-3}$ with batch size 128, or $1{,}024$ above $20{,}000$ points, for up to 300 epochs and we stopped once the error on a tenth of the training points did not improve for 20 epochs. Both estimate the mean of the points at a position, which is the best prediction from the position in squared error, and the mean of $k$ points at the position exceeds its error by a share of $1/k$. The mean of the 15 training points nearest to a held-out point in the input uses no map and shows what the positions lose.

\paragraph{Errors on every data set.} FloDR reached $0.55$ and $0.33$ on MNIST and Fashion-MNIST, versus $0.90$ and $0.60$ for the inverse transform of UMAP, and $0.64$ and $0.35$ for the decoder of Parametric UMAP. On paul15, where relatively little can be read from a position, the three reached $0.93$, $1.00$, and $0.96$, and on the three atlases $0.28$, $0.48$, and $0.62$ (FloDR), $0.32$, $0.52$, and $0.66$ (UMAP), and $0.43$, $0.68$, and $0.72$ (Parametric UMAP). The mean of the 15 nearest training points on the map of UMAP reached $0.55$, $0.30$, $1.00$, $0.31$, $0.52$ and $0.67$ on the six data sets. This means that on the images most of FloDR's lead is over the transform, not the map, of UMAP. On the map of FloDR, the same mean was lower on every split, by $0.001$ to $0.08$, and a decoder trained on its layout was more accurate than the decode on four of the six data sets by at most $0.03$. At the position of the forward pass instead of the optimised step, the error of FloDR was $0.84$ and $0.56$ on MNIST and Fashion-MNIST, $0.96$ on paul15, and $0.39$, $0.64$, and $0.81$ on the atlases. It was above that of Parametric UMAP on 19 of the 24 splits, and above that of UMAP on every split of the atlases and on no split of the other data sets.

\paragraph{Draws from the density.} On one split with $20$~\% of the cells held out and three fits, we drew 64 residuals from the fitted density at the forward-pass position of a held-out cell and decoded them. The range between the 5th and the 95th percentile of the draws was $3.6$ to $3.7$ times that of the 30 training cells nearest to the position on paul15, and $26.8$ to $30.6$ times on the bone marrow atlas (median over the 50 components and up to $2{,}000$ held-out cells). The coverage certificate refused all three of the bone marrow fits and was undecided on the three of paul15.

\section*{Supplementary Note 2: Tests against unseen measurements}

\paragraph{Maps.} In six of the seven tests, FloDR was compared with UMAP, openTSNE, PaCMAP, DREAMS, PHATE, and the first two principal components, with densMAP~\cite{Narayan2021} (settings of UMAP) in the three lineage tests and the protein test, and with Parametric UMAP in the genotype and cortex tests. FloDR without the ordinal term was a control on LARRY, the protein data, the genotypes, and the cortex. In the seventh test, the placement of progenitors, FloDR was compared with UMAP, openTSNE, PHATE, and Parametric UMAP, next to its own forward pass and two transfers from the nearest fitted cells (Supplementary Table~4g,h).

\paragraph{Lineage.} LARRY~\cite{Weinreb2020} ($130{,}887$ cells) and the LSK (lineage-negative, Sca-1-positive, c-Kit-positive) progenitor and reprogramming arms of CellTag-multi~\cite{Jindal2023} ($61{,}662$ and $232{,}337$ cells) were reduced to $2{,}000$ variable genes of log-normalised, scaled counts. Fate is the AUROC for ranking barcoded early cells by the majority fate of their clone, from the difference of their map distances to the two late fate medians. Coupling is the Spearman correlation, over pairs of late cell types, of their normalised covariance within clones with the negative median map distance. Sisters is whether sister cells lie closer on the map than matched non-sisters. For the placement test, early cells of LARRY and of the LSK arm were placed on maps of the late cells, and fate was read from the placed cells.

\paragraph{Protein levels.} In 12 CITE-seq batches~\cite{luecken2021sandbox} ($90{,}261$ cells), map distances were compared with distances in 134 protein levels between cell types and between cells.

\paragraph{Genotypes.} From the HGDP-CEPH genomes~\cite{Bergstrm2020}, we kept autosomal biallelic SNPs (minor allele frequency $\geq0.05$, missingness $\leq0.02$), pruned for linkage ($r^2<0.2$) to $212{,}965$, removed $11$ of $929$ individuals as relatives, and standardised genotypes per SNP. Principal components were fitted on 47 populations ($795$ individuals). Seven further populations ($123$ individuals), in each region the one of median size, were held out, projected onto them, and placed by the out-of-sample step of each method. The readout is the mean over populations of the Spearman correlation of map distances between population medians with great-circle distances between sampling sites through five waypoints~\cite{Ramachandran2005}. In Fig.~5a,b, each map is rotated and reflected, without scaling, to the best match of its $795$ individuals with their sampling coordinates, the Americas counted east of Oceania.

\paragraph{Cortical depth.} We used $139{,}380$ glutamatergic neurons of SEA-AD MERFISH data~\cite{Gabitto2024}, 19 donors mapped and 8 placed, and per donor the Spearman correlation of map distances between ten depth bins with their separation.

\end{document}